\documentclass{article}
\usepackage[accepted]{icml2021}
\usepackage{graphicx}
\usepackage[ruled,vlined]{algorithm2e}
\usepackage{amssymb}
\usepackage{amsmath}
\usepackage{amsthm}
\usepackage{dsfont}
\usepackage{multicol}
\usepackage{xcolor}
\usepackage{parskip}
\usepackage{color,soul}
\usepackage{subfigure}

\setul{0.5ex}{0.3ex}
\setulcolor{red}

\usepackage{natbib}
\usepackage{tikz}
\usepackage{longtable}
\usepackage{booktabs}

\usepackage[tableposition=below]{caption}
\usepackage{pifont}
\newcommand*\numcircledtikz[1]{\tikz[baseline=(char.base)]{
            \node[shape=circle,draw, inner sep=.1ex] (char) {#1};}} 

\newcommand{\E}{\mathbf{E}}
\newcommand{\KL}{\text{KL}}
\newcommand{\Tr}{\text{Tr}}
\newcommand{\R}{\mathbb{R}}

\newtheorem{thm}{Theorem}[section]

\newtheorem{lem}[thm]{Lemma}

\theoremstyle{definition}

\renewcommand*{\vec}[1]{\boldsymbol{#1}}

\DeclareMathOperator*{\argmin}{arg\,min}

\icmltitlerunning{Generative Particle Variational Inference via Estimation of Functional Gradients}

\begin{document}
\twocolumn[

\icmltitle{Generative Particle Variational Inference via Estimation of Functional Gradients}
\icmlsetsymbol{equal}{*}

\begin{icmlauthorlist}
\icmlauthor{Neale Ratzlaff}{equal,osu}
\icmlauthor{Qinxun Bai}{equal,horizon}
\icmlauthor{Li Fuxin}{osu}
\icmlauthor{Wei Xu}{horizon}
\end{icmlauthorlist}

\icmlaffiliation{osu}{Department of Electrical Engineering and Computer Science, Oregon State University, Corvallis, Oregon}
\icmlaffiliation{horizon}{Horizon Robotics, Cupertino, California}
\icmlcorrespondingauthor{Neale Ratzlaff}{ratzlafn@oregonstate.edu}
\icmlkeywords{Bayesian Deep Learning}
\vskip 0.3in
]
\printAffiliationsAndNotice{\icmlEqualContribution} 

\begin{abstract}  
   Recently, particle-based variational inference (ParVI) methods have gained interest because they can avoid arbitrary parametric assumptions that are common in variational inference. However, many ParVI approaches do not allow arbitrary sampling from the posterior, and the few that do allow such sampling suffer from suboptimality.
   This work proposes a new method for learning to approximately sample from the posterior distribution. 
   We construct a neural sampler that is trained with the functional gradient of the KL-divergence between the empirical sampling distribution and the target distribution, assuming the gradient resides within a reproducing kernel Hilbert space. 
   Our generative ParVI (GPVI) approach maintains the asymptotic performance of ParVI methods while offering the flexibility of a generative sampler. 
   Through carefully constructed experiments, we show that GPVI outperforms previous generative ParVI methods such as amortized SVGD, and is competitive with ParVI as well as gold-standard approaches like Hamiltonian Monte Carlo for fitting both exactly known and intractable target distributions. 
\end{abstract}

\section{Introduction}

Bayesian inference provides a powerful framework for reasoning and prediction under uncertainty. However, computing the posterior is tractable with only a few parametric distributions, making wider applications of Bayesian inference difficult. Traditionally, MCMC and variational inference methods are utilized to provide tractable approximate inference, but these approaches face their own difficulties if the dimensionality of the space is extremely high. For example, a recent case of interest is Bayesian neural networks (BNNs), which applies Bayesian inference to deep neural network training in order to provide a principled way to assess model uncertainty. The goal in this regime is to model the posterior of every parameter in all the weight tensors from every layer of a deep network. However, developing efficient computational techniques for approximating this intractable posterior with extremely high dimensionality remains challenging.

Recently, particle-based variational inference (ParVI) methods \citep{liu2016stein, liu2019understanding, liu2017GF} have been proposed to represent the variational distribution by a set of particles and update them through a deterministic optimization process to approximate the posterior. While achieving both asymptotic accuracy and computational efficiency, ParVI methods are restricted by the fixed number of particles and lack the ability of drawing new samples beyond the initial set of particles. To address this issue, amortized ParVI methods \cite{wang2016learning} have been proposed to amortize the ParVI gradients in training a neural sampler. While being flexible in drawing samples,  in Sec. 4 we show that amortized ParVI methods cannot match the convergence behavior of ParVI methods.

\begin{figure*}[ht!]
\centering  
    \subfigure[HMC]{
        \label{fig:hmc_regression}
        \vspace{-1cm}
        \includegraphics[width=.23\linewidth]{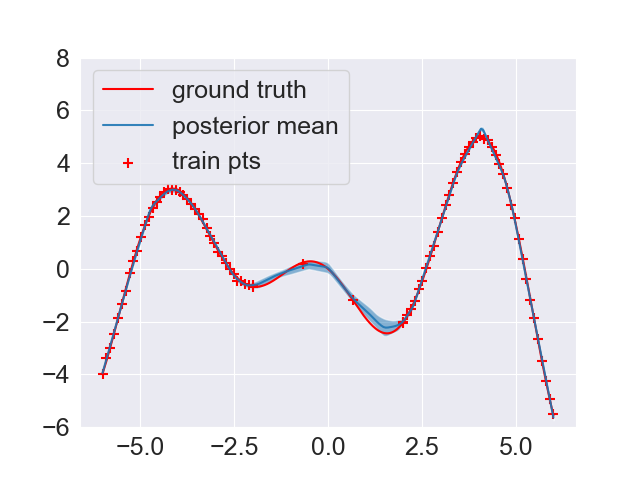}
    }
    \subfigure[SVGD]{
        \label{fig:hmc_regression}
        \vspace{-1cm}
        \includegraphics[width=.23\linewidth]{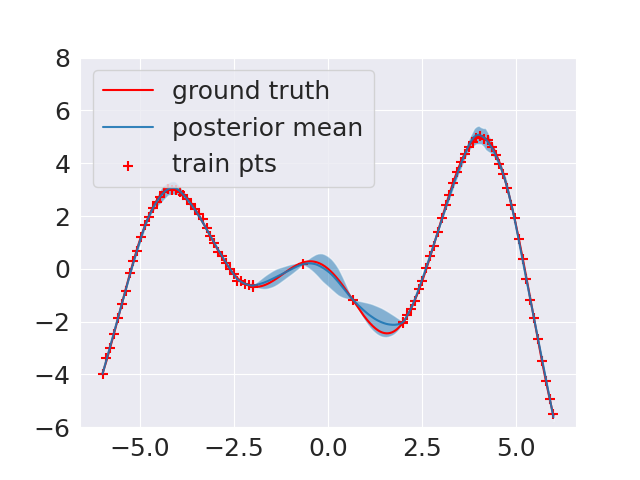}
    }
    \subfigure[GPVI]{
         \label{fig:hmc_regression}
         \includegraphics[width=.23\linewidth]{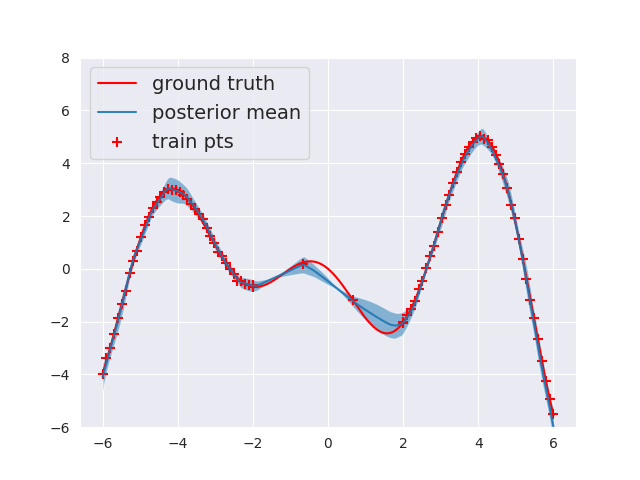}
     }
    \subfigure[Amortized SVGD]{
        \label{fig:asvgd_regression}
        \includegraphics[width=.23\linewidth]{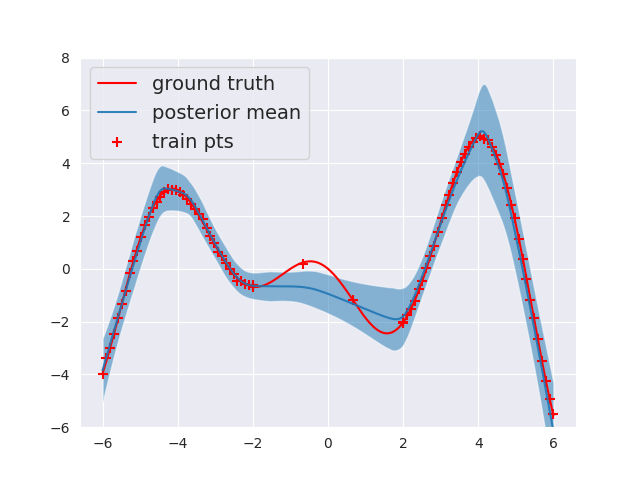} 
    }
\vskip -0.1in
\caption{Predictive uncertainty of methods for a 1-D regression task. 
(a) HMC predictive posterior matches the uncertainty in the data; 
(b) SVGD performs comparably to HMC; 
(c) Our proposed GPVI performs similarly to SVGD, with the additional capability of sampling  new particles during inference;
(d) Amortized SVGD overestimates the uncertainty when the data is sparse.} 
\label{fig:1d-regression}
\vskip -0.05in
\end{figure*}

In this work, we propose a generative particle variational inference (GPVI) approach that addresses those issues.
GPVI trains a neural sampler network by directly estimating the functional gradient w.r.t. the KL-divergence between the distribution of generated particles and the target distribution, and pulls it back to update the neural sampler. As such it allows the neural sampler to directly generate particles that match the posterior distribution, hence achieving the asymptotic accuracy and computational efficiency of ParVI methods. In figure \ref{fig:1d-regression}, we show that the predictive distribution of 1D regression functions sampled from GPVI nearly matches that from the ParVI solution, while amortized ParVI fails.

The main computational challenge lies in a reliable estimate of the functional gradient that involves the inverse of the input-output Jacobian of the neural sampler. Instead of directly computing this term and paying a high computational cost, we introduce a helper network to estimate the inverse Jacobian vector product and train the helper network via gradient descent. By alternating between this gradient step and the gradient update of the sampler network, the computational cost is distributed over the whole training procedure.

In experiments, our proposed approach achieves comparable convergence performance as ParVI methods. It is considerably superior than that of amortized ParVI methods, while still allowing efficient sampling from the posterior. By directly applying our approach as a hypernetwork to generate BNNs, we achieve competitive performance regarding uncertainty estimation.

In summary, our contributions are three-fold,
\begin{itemize}
\vspace{-0.05in}
    \item We propose GPVI, a new variational inference approach that trains a neural sampler to generate particles from any posterior distribution. GPVI estimates the functional gradient and uses it to update the neural sampler. It enjoys the asymptotic accuracy and computational efficiency of ParVI methods. Comparing with existing amortized ParVI methods, our approach enjoys the same efficiency and flexibility while showing considerable advantage in convergence behavior.
\vspace{-0.03in}
    \item We design careful techniques for efficient gradient estimates that address the challenges in approximating the product between the inverse of the Jacobian and a vector. 
\vspace{-0.03in}
    \item We apply our approach to BNNs and achieve competitive uncertainty estimation quality for deep neural networks.
    
\vspace{-0.05in}
\end{itemize}

\section{Related Work}

ParVI is a recent class of non-parametric variational inference methods.  \cite{liu2016stein,liu2017GF} proposed stein variational gradient descent (SVGD), which  deterministically updates an empirical distribution of particles toward the target distribution via a series of sequentially constructed smooth maps. Each map defines a perturbation of the particles in the direction of steepest descent towards the target distribution under the KL-divergence metric. 
\citet{liu2019understanding} cast ParVI as a gradient flow on the space of probability measures $\mathcal{P}_2$ equipped with the $W_2$ Wasserstein metric, and proposed 
ParVI methods GFSD and GFSF by smoothing the density and the function respectively.
A potential limitation of SVGD and GFSF is the restriction of the transport maps to an RKHS. The fisher neural sampler \cite{hu2018stein, grathwohl2020learning} lifts the function space of transportation maps to the space of $\mathcal{L}_2$ functions, and performs a minimax optimization procedure to find the optimal perturbation. As a non-parametric approach, ParVI does not depend on the specific parameterization of the approximate posterior. 
ParVI is also more efficient than MCMC methods due to its deterministic updates and gradient-based optimization. 

The limitation of ParVI is the inability to draw additional samples beyond the initial set. To address this limitation, our approach allows for arbitrary draws of new samples once the generator is trained.
The most notable generative ParVI approach prior to our work was amortized SVGD \cite{wang2016learning}, which
proposed using a generative model to approximately sample from the SVGD particle distribution. Amortized SVGD performs a one-step update that backpropagates the SVGD gradient back to the generator parameters, which is often not optimal. 
A more detailed discussion of the differences between our method and GPVI is presented in Sec.~\ref{sec:amortized_par_vi}.

Performing direct Bayesian inference in the weight-space of neural networks \cite{neal2012bayesian, buntine1991bayesian, mackay1992practical} is known to be intractable in the context of deep learning. Thus, approximations have been formulated using Langevin Dynamics \cite{welling2011bayesian, korattikara2015bayesian}, Monte Carlo dropout \cite{gal2016dropout, kingma2015variational}, and variational inference with mean-field Gaussian priors \cite{graves2011practical}. 
The variational approximation has been iterated upon extensively. Bayes by Backprop \cite{blundell2015weight} proposed an unbiased gradient estimator that allowed for training of deep Bayesian neural networks using the local reparameterization trick from \citet{kingma2015variational}.
However, fully factorized Gaussian approximations are insufficient to capture the complex structure of a high dimensional posterior  \cite{yao2019quality}. 
As a result, structured approximate posteriors have been proposed using matrix Gaussians \cite{louizos2016structured}, normalizing flows \cite{louizos2017multiplicative, krueger2017bayesian}, and hypernetworks \cite{pawlowski2017implicit}.

Evaluating the performance of BNNs is difficult, as there is often no access to the ground truth posterior. Therefore, prior work often considers a ``gold-standard" method such as HMC \cite{neal2011mcmc}, or exact Gaussian processes \cite{rasmussen2003gaussian} as a proxy for the true posterior. While these methods set a high bar for BNNs, they are in general not scalable to large-scale problems. Nonetheless there were  attempts to bring HMC \cite{strathmann2015gradient, chen2014stochastic} and GPs \cite{wilson2016stochastic, cheng2017variational, wang2019exact, hensman2013gaussian, hensman2015scalable} in the realm of tractability for high dimensional problems.  

Variational inference via smooth transport maps is also used in normalizing flows \citep{rezende2015variational}. But the transport maps used in normalizing flows and in \citet{marzouk2016introduction} are bijections which makes them unsuitable for learning distributions over neural network functions.

\section{Generative Particle Variational Inference}
\label{sec:approach}

We want to learn a parametric generator: $\vec{f}_{\vec\theta}: \mathcal{Z}\to\R^d$, parameterized by
$\vec\theta$,
where $\mathcal{Z}\subset\R^d$ is the convex space of input noise and
$\vec{x}=\vec{f}_{\vec\theta}(\vec{z})$ generates a sample $\vec{x}$ from an input noise $\vec{z}$.
Let $q(\vec{x})$ represents the implicit distribution of samples generated by $\vec{f}_{\vec\theta}(\vec{z})$, where $\vec{z}\sim N(\vec{0}, I_d)$. Let $p(\vec{x})$ be the target distribution, we want to solve for $\vec{f}_{\vec\theta}$ that minimizes the objective $\KL\left(q(\vec{x})\|p(\vec{x}\right))$. 

Our approach treats this problem from a functional optimization perspective, by first computing the functional gradient of the objective, and then pulling it back to parameter space through the function parameterization. In the rest of this section we introduce our algorithm in detail and compare it with amortized ParVI approaches.

\subsection{Functional gradient and its pullback}
\label{sec:func_grad}

Let $\mathcal{J}(\vec{f})=\KL\left(q(\vec{x})\|p(\vec{x})\right)$ be the objective, where $\vec{x}=\vec{f}(\vec{z})$.
If $\vec{f}$ is injective, by change of variables for probability measure, 
\begin{equation}
\label{eq:dist_trans}
q(\vec{x})=\frac{p_{\vec{z}}(\vec{z})} {\left|\det\left(\frac{\partial\vec{f}}{\partial\vec{z}}\right)\right|},
\end{equation}
where $p_{\vec{z}}(\vec{z})$ is the distribution from which $\vec{z}$ is sampled. The minimization objective becomes,

\begin{equation}
    \label{eq:f_obj}
    \mathcal{J}(\vec{f}) = \mathbf{E}_{\vec{z}}\left[-\log p(\vec{f}(\vec{z})) 
    + \log\frac{p_{\vec{z}}(\vec{z})}
        {\left|\det\left(\frac{\partial\vec{f}}{\partial\vec{z}}\right)\right|} \right].
\end{equation}

Consider some function approximation of $\vec{f}$, say $\vec{f}=\vec{f}_{\vec\theta}$, then the minimization objective becomes,
\begin{equation}
    \label{eq:theta_obj}
    \mathcal{J}(\vec\theta) = \mathbf{E}_{\vec{z}}\left[-\log p(\vec{f}_{\vec\theta}(\vec{z})) 
    + \log\frac{p_{\vec{z}}(\vec{z})}
        {\left|\det\left(\frac{\partial\vec{f}_{\vec\theta}}{\partial\vec{z}}\right)\right|}\right].
\end{equation}
Directly computing the gradient of $\mathcal{J}(\vec\theta)$ w.r.t. $\vec\theta$ involves not only an inverse of the Jacobian $\left(\frac{ \partial\vec{f} }{ \partial\vec{z}} \right)^{-1}$, but also second derivatives of $\vec{f}_{\vec\theta}$ (see Appendix A for details), which is overly expensive to compute in practice. 

We propose, instead, to first compute the functional gradient of~\eqref{eq:f_obj} w.r.t. $\vec{f}$, i.e., $\nabla_{\vec{f}}\mathcal{J}(\vec{f})$, and then 
back-propagate it through the generator to get the gradient w.r.t. $\vec\theta$, i.e.,
\begin{equation}
\label{eq:rkhs_theta_grad}
\nabla_{\vec\theta}\mathcal{J}=\mathbf{E}_{\vec{z}}\left[
\frac{\partial \vec{f}(\vec{z})}{\partial\vec\theta}
\nabla_{\vec{f}}\mathcal{J}(\vec{f})(\vec{z}) \right].
\end{equation}

The following theorem gives an explicit formula for computing the functional gradient 
$\nabla_{\vec{f}}\mathcal{J}(\vec{f})$ when $\vec{f}$ is chosen from a Reproducing kernel Hilbert space (RKHS). The proof is provided in Appendix A.
\begin{thm}
\label{thm:rkhs_func_grad}
Let $\vec{x}=\vec{f}(\vec{z})$, where $\vec{z}\sim p_{\vec{z}}(\vec{z})$, vector function $\vec{f}=(f^1, \ldots, f^d)\in\mathcal{H}^d$ with $f^i\in\mathcal{H}$, where $\mathcal{H}$ is the RKHS with kernel $k(\cdot, \cdot)$, $\mathcal{H}^d$ is equipped with inner product $\langle \vec{f}, \vec{g}\rangle_{\mathcal{H}^d}=\sum^d_{i=1}\langle f^i, g^i\rangle_{\mathcal{H}}$. For $\mathcal{J}(\vec{f})$ well-defined by~\eqref{eq:f_obj}, we have 
\begin{equation}
\label{eq:rkhs_func_grad}
\begin{aligned}
    \nabla_{\vec{f}}\mathcal{J}(\vec{f})(\vec{z})
    =\ & \mathbf{E}_{\vec{z}'}\bigg[ -\nabla_{\vec{x}}\log p(\vec{x})
        \bigg|_{\vec{x}=\vec{f}(\vec{z}')}
    k(\vec{z}', \vec{z}) \\
    &- \left(\frac{ \partial\vec{f} }{ \partial\vec{z}' } \right)^{-1}
    \nabla_{\vec{z}'}k(\vec{z}', \vec{z})
    \bigg].
\end{aligned}
\end{equation}
\end{thm}

\subsection{Reparameterization of $\vec{f}$}
There are two considerations for reparameterizing $\vec{f}$. Firstly, in order for the inverse $\left(\frac{ \partial\vec{f} }{ \partial\vec{z}'}\right)^{-1}$ in~\eqref{eq:rkhs_func_grad} to be well-defined, the Jacobian $\frac{ \partial\vec{f} }{ \partial\vec{z}' }$ should be a square matrix, i.e., input noise $\vec{z}$ should have the same dimension as the output $\vec{x}=\vec{f}(\vec{z})$. In practice, especially those applications involving BNNs, each $\vec{x}$ represents parameters of a sampled neural network and therefore can be extremely high dimensional. As a result, a high dimensional $\vec{f}$ can be computationally prohibitive. 
Secondly, in order for the change of variables density formula~\eqref{eq:dist_trans} to hold, $\vec{f}$ needs to be injective, which is in general not guaranteed for an arbitrary neural network function. 

To overcome the above two concerns, we consider the following parameterization,
\begin{equation}
\label{eq:generator}
    \vec{f}_{\vec\theta}(\vec{z}) = \vec{g}_{\vec\theta}\left(\vec{z}^{(:k)}\right) + \lambda \vec{z}, \quad \forall\vec{z}\in\R^d,
\end{equation}
where $\vec{z}^{(:k)}\in\R^k$ denotes the vector consisting of the first $k$ components of $\vec{z}$,
and $\vec{g}_{\vec\theta}: \R^k\to\R^d$ with parameters $\vec\theta$ is a much slimmer neural network.
In our experiments, $\vec{g}_{\vec\theta}$ is designed with an input dimension $k$ less than $30\%$ the size of $d$. For high dimensional open-category experiments where $d > 60,000$, we use a $k$ of less than $2\%$ of $d$. 
For $\vec{f}_{\vec\theta}$ defined by~\eqref{eq:generator}), the Jacobian is,
\begin{equation}
\label{eq:jac_gen}
\left[\frac{\partial\vec{f}_{\vec\theta}}{\partial\vec{z}}\right]_{d\times d} =     
    \left[ \left[\frac{\partial\vec{g}_{\vec\theta}}{\partial\vec{z}^{(:k)}}\right]_{d\times k} 
        \bigg| \vec{0}_{d\times (d-k)} \right]_{d\times d}
    + \lambda \vec{I}_d,
\end{equation}
where $\lambda$ is a hyper-parameter. 
Note that for sufficiently large $\lambda$, the Jacobian defined by~\eqref{eq:jac_gen} is positive definite. 
Since the domain $\mathcal{Z}$ of $\vec{z}$ is convex, it is straightforward to show that $\vec{f}_{\vec\theta}$ defined by~\eqref{eq:generator} is injective. We include the proof in Appendix A.

In practice we set $\lambda$ to be $1.0$ and find it sufficient throughout our experiments. 

\subsection{Estimating the Jacobian inverse}
\label{sec:p_inverse}

The main computational challenge of~\eqref{eq:rkhs_func_grad} lies in computing the term
\begin{equation}
\label{eq:lin_sys}
    \left(J_{\vec{f}}(\vec{z}')\right)^{-1} \nabla_{\vec{z}'}k(\vec{z}', \vec{z}),
\end{equation}
 where $J_{\vec{f}}(\vec{z}')=\frac{ \partial\vec{f} }{ \partial\vec{z}' }, $ especially considering that we need an efficient implementation for batched $\vec{z}$ and $\vec{z}'$. 

Directly evaluating and storing the full Jacobian $J_{\vec{f}}(\vec{z}')$ for each $\vec{z}'$ of the sampled batch is not acceptable from the standpoint of either time or memory consumption. There exist iterative methods for solving the linear equation system
$\vec{y}=\left(J_{\vec{f}}(\vec{z}')\right)^{-1} \nabla_{\vec{z}'}k(\vec{z}', \vec{z})$ \citep{young1954iterative, fletcher1976conjugate}, 
which involves computing the vector-Jacobian product
$J_{\vec{f}}(\vec{z}') \nabla_{\vec{z}'}k(\vec{z}', \vec{z})$ 
at each iteration. By alternating between this iterative solver and the gradient update~\eqref{eq:rkhs_theta_grad}, it is possible to get a computationally amenable algorithm. However, as shown in the Appendix C, such an algorithm does not converge to the target distribution $p$ even for a simple Bayesian linear regression task. This is due to the fact that batches of both $\vec{z}$ and $\vec{z}'$ for evaluating~\eqref{eq:rkhs_func_grad} need to be re-sampled for each gradient update~\eqref{eq:rkhs_theta_grad} to avoid the cumulative sampling error. Therefore, the above alternating procedure of iterative solver for 
$\vec{y}=\left(J_{\vec{f}}(\vec{z}')\right)^{-1} \nabla_{\vec{z}'}k(\vec{z}', \vec{z})$
ends up shooting a moving target for different batches of $\vec{z}$ and $\vec{z}'$ at each iterate, which is difficult for  convergence.

To overcome this computational challenge, we propose a helper network, denoted by 
$\vec{h}_{\vec{\eta}}(\vec{z}', \nabla_{\vec{z}'}k)$, and parameterized by $\vec\eta$,
that consumes both $\vec{z}'$ and $\nabla_{\vec{z}'}k(\vec{z}', \vec{z})$ and predicts
$\left(J_{\vec{f}}(\vec{z}')\right)^{-1} \nabla_{\vec{z}'}k(\vec{z}', \vec{z})$. 
With the helper network, the functional gradient~\eqref{eq:rkhs_func_grad} can be computed by,
\begin{equation}
\label{eq:rkhs_func_grad_helper}
\begin{aligned}
    \nabla_{\vec{f}}\mathcal{J}(\vec{f})(\vec{z})
    =\ & \mathbf{E}_{\vec{z}'}\bigg[ -\nabla_{\vec{x}}\log p(\vec{x})
        \bigg|_{\vec{x}=\vec{f}(\vec{z}')}
    k(\vec{z}', \vec{z}) \\
    &- \vec{h}_{\vec\eta}\left(\vec{z}', \nabla_{\vec{z}'}k(\vec{z}', \vec{z}) \right)
    \bigg].
\end{aligned}
\end{equation}
We use the following loss to train the helper network,
\begin{equation}
\label{eq:loss_helper}
    \mathcal{L}(\vec\eta) = \|J_{\vec{f}}(\vec{z}')\vec{h}_{\vec\eta}
        (\vec{z}', \nabla_{\vec{z}'}k(\vec{z}', \vec{z}))
    - \nabla_{\vec{z}'}k(\vec{z}', \vec{z})\|^2,
\end{equation}
where $J_{\vec{f}}(\vec{z}')\vec{h}_{\vec\eta}$ can be computed by the following formula given the reparameterization~\eqref{eq:generator} of $\vec{f}_{\vec\theta}$,
\begin{equation}
\label{eq:vjp_gen}
\vec{h}^T_{\vec\eta}\left(\frac{\partial\vec{f}_{\vec\theta}}{\partial\vec{z}'}\right) =     
    \vec{h}^T_{\vec\eta}\left(\frac{\partial\vec{g}_{\vec\theta}}{\partial\vec{z}'^{(:k)}}\right) 
    + \lambda \vec{h}^T_{\vec\eta},
\end{equation}
where the Vector-Jacobian Product (VJP)
$\vec{h}^T_{\vec\eta}\left(\frac{\partial\vec{g}_{\vec\theta}}{\partial\vec{z}'^{(:k)}}\right)$
can be computed by one backward pass of the function $\vec{g}_{\vec\theta}$. 

\subsection{Summary of the algorithm}
\label{sec:alg_summary}

Both $\vec{g}_{\vec\theta}$ and $\vec{h}_{\vec{\eta}}$ can be trained with stochastic gradient descent (SGD), and our algorithm alternates between the SGD updates of $\vec{g}_{\vec\theta}$ and $\vec{h}_{\vec{\eta}}$. Note that the helper network $\vec{h}_{\vec{\eta}}$ only has to chase the update of $\vec{g}_{\vec\theta}$, but no more extra moving targets due to re-sampling of $\vec{z}$ and $\vec{z}'$. Our experiments in Sec.~\ref{sec:exp} and the appendix show that the helper network is able to efficiently approximate~\eqref{eq:lin_sys} which enables the convergence of the gradient update~\eqref{eq:rkhs_theta_grad}.
Our overall Generative Particle VI (GPVI) algorithm is summarized in Algorithm~\ref{alg:gpvi}.

\vspace{-0.05in}
\begin{algorithm}
    \SetAlgoLined
    \textbf{Initialize} generator $\vec{g}_{\vec\theta}$, helper $\vec{h}_{\eta}$, and 
        learning rate $\epsilon$ \\

    \While{\normalfont{Not converge}}{
        1. sample two batches $\{\vec{z}_i\}, \{\vec{z}'_i\}\sim N(\vec{0}, I_d)$ \\
        2. compute $k(\vec{z}', \vec{z})$ and $\nabla_{\vec{z}'}k(\vec{z}', \vec{z})$ \\
        3. forward $\vec{g}_{\vec\theta}$ to compute $\vec{f}_{\vec\theta}(\vec{z})$
            and $\vec{f}_{\vec\theta}(\vec{z}')$ by~\eqref{eq:generator} \\
        4. forward $\vec{h}_{\vec\eta}$ to compute 
            $\vec{h}_{\vec\eta}(\vec{z}', \nabla_{\vec{z}'}k(\vec{z}', \vec{z}))$ \\
        5. backward $\vec{g}_{\vec\theta}$ to compute the VJP
            $\vec{h}^T_{\vec\eta}\left(\frac{\partial\vec{g}_{\vec\theta}}{\partial\vec{z}'^{(:k)}}\right)$ \\
            \phantom{lor} and then construct $J_{\vec{f}}(\vec{z}')\vec{h}_{\vec\eta}$ by~\eqref{eq:vjp_gen},\\
        6. update $\vec{h}_{\vec\eta}$ by 
            $\vec\eta\leftarrow\vec\eta - \epsilon\nabla_{\vec\eta}\mathcal{L}$, \\
            \phantom{lor} where $\nabla_{\vec\eta}\mathcal{L}$ is computed by 
            back-propagating~\eqref{eq:loss_helper} \\
        7. compute the functional gradient by~\eqref{eq:rkhs_func_grad_helper} \\
        8. update $\vec\theta$ by 
            $\vec\theta\leftarrow\vec\theta - \epsilon\nabla_{\vec\theta}\mathcal{J}$, \\
            \phantom{lor} where $\nabla_{\vec\theta}\mathcal{J}$ is computed 
            by~\eqref{eq:rkhs_theta_grad} 
     }
    \caption{Generative Particle VI (GPVI)}
    \label{alg:gpvi}
\end{algorithm}
\vspace{-0.15in}

\subsection{Comparison with Amortized SVGD}
\label{sec:amortized_par_vi}

Stein variational gradient descent (SVGD) represents $q(\vec{x})$ by a set of particles $\{\vec{x}_i\}^n_{i=1}$, which are updated iteratively by,
\begin{equation}
\label{eq:par_perturb}
\vec{x}_i \leftarrow \vec{x}_i + \epsilon\vec\phi^*(\vec{x_i}),
\end{equation}
where $\epsilon$ is a step size and $\vec\phi^*: \R^d \to \R^d$ is a vector field (perturbation) on the space of particles that corresponds to the optimal direction to perturb particles, i.e.,
\begin{equation*}
    \vec\phi^* = \argmin_{\vec\phi\in\mathcal{F}}
        \left\{ \frac{d}{d\epsilon}\KL(q_{[\epsilon\vec\phi]}(\vec{x})\|p(\vec{x}))\bigg|_{\epsilon=0}
        \right\},
\end{equation*}
where $q_{[\epsilon\vec\phi]}(\vec{x})$ denotes the density of particles updated by~\eqref{eq:par_perturb} using the perturbation $\vec\phi$, where the density of original particles is $q(x)$. When $\mathcal{F}$ is chosen to be the unit ball of some RKHS $\mathcal{H}$ with kernel function $k(\cdot, \cdot)$, SVGD gives the following closed form solution for $\vec\phi^*$,
\begin{equation}
\label{eq:svgd_grad}
\vec\phi^*(\vec{x}) = \mathbf{E}_{\vec{x}'\sim q}\left[\nabla_{\vec{x}'}\log p(\vec{x}')k(\vec{x}', \vec{x})
    + \nabla_{\vec{x}'}k(\vec{x}', \vec{x}) \right].
\end{equation}

To turn SVGD into a neural sampler,
Amortized SVGD~\cite{wang2016learning} first samples particles from a generator and then back-propagates the particle gradients~\eqref{eq:svgd_grad} through the generator to update the generator parameters. 
Let $\vec{x}=\vec{f}_{\vec\theta}(\vec{z}),\ \vec{z}\sim N(\vec{0}, I_d)$ be the particle generating process, where $\vec{f}_{\vec\theta}$ is the generator parameterized by $\vec\theta$, amortized SVGD updates $\vec\theta$ by,
\begin{equation}
\label{eq:asvgd_update}
\vec\theta \leftarrow \vec\theta 
    + \epsilon\sum_{i=1}^m \frac{\partial\vec{f}_{\vec\theta}(\vec{z}_i)}{\partial\vec\theta}
        \vec\phi^*(\vec{f}_{\vec\theta}(\vec{z}_i)),
\end{equation}
where $\vec\phi^*(\vec{f}_{\vec\theta}(\vec{z}_i))$ is computed by~\eqref{eq:svgd_grad}.

The following Lemma gives an explicit view regarding what functional gradient amortized SVGD back-propagates through the generator. The proof is given in the Appendix A.

\begin{lem}
\label{lem:svgd_func_grad}
If particles are generated by $\vec{x}=\vec{f}(\vec{z}),\ \vec{z}\sim p_{\vec{z}}(\vec{z})$,
Eqn.~\eqref{eq:svgd_grad} is the functional gradient of the KL objective w.r.t. the perturbation function $\vec\phi: \R^d\to\R^d$ applied on the output space of $\vec f$, 
i.e., 

\begin{equation}
\label{eq:svgd_func_loss}
    \nabla_{\vec\phi}\bigg|_{\vec\phi=\text{id}}\KL\left(q(\vec{x})\|p(\vec{x})\right)= -\vec\phi^*,
\end{equation}
where $\vec{x}=\vec\phi(\vec{f}(\vec{z}))$
and $\vec\phi=(\phi^1, \ldots, \phi^d)\in\mathcal{H}^d$ with $\phi^i\in\mathcal{H}$, where $\mathcal{H}$ is the RKHS with kernel $k(\cdot, \cdot)$, $\mathcal{H}^d$ is equipped with inner product $\langle \vec\phi, \vec\xi\rangle_{\mathcal{H}^d}=\sum^d_{i=1}\langle \phi^i, \xi^i\rangle_{\mathcal{H}}$.
\end{lem}

To see the difference between our GPVI and amortized SVGD, both methods learn a generator $\vec f$ that generates particles $\vec x$ from input noise $\vec z$. To update $\vec f$, gradients of the KL objective $\mathcal{J}(\vec{f})$ have to be backpropagated to $\vec f$ in both cases. Here, the best is to use the steepest descent direction for $\vec f$, which GPVI does (proved in Thm~\ref{thm:rkhs_func_grad}). As Lemma~\ref{lem:svgd_func_grad} shows, the SVGD 
gradient $\vec\phi^*$ given by~\eqref{eq:svgd_grad}
is optimal for a transformation $\vec\phi$ applied to the set of current particles. In in both amortized SVGD and GPVI, such a set is not kept 
and $\vec\phi$ does not exist since they only maintain and update $\vec f$.
But amortized SVGD directly uses $\vec\phi^*(\vec f)$ in~\eqref{eq:asvgd_update} as the direction
to update $\vec f$,  which is an unproven use -- it only coincides with the steepest descent direction for $\vec f$ used by GPVI in special cases such as $\vec f = \text{id}$, where~\eqref{eq:rkhs_func_grad} and~\eqref{eq:svgd_grad} are equivalent because $\vec{z} = \vec{x}$ and the Jacobian is identity. 
Regarding the definition of RKHS, GPVI's RKHS approximates the tangent space of $\vec{f}$, it is therefore naturally defined on $\vec{z}$. 
(Amortized) SVGD uses RKHS to approximate the tangent space of $\vec\phi$, which is naturally defined on $\vec x$.

The amortizing step applied in both methods, i.e., back-propagating some functional gradient to update the generator parameters, is in general not guaranteed to keep the original descent direction of the functional gradient. However, GPVI amortizes the steepest descent direction which is optimal in first order sense, while amortized SVGD amortizes a non-steepest descent direction, which is more likely to result in a non-descending direction after amortizing.
In practice, as shown in the next section, our approach consistently outperforms amortized SVGD in approximating the target distribution and capturing model uncertainty.

\section{Experiments}
\label{sec:exp}
To demonstrate the effectiveness of our approach for approximate Bayesian inference, we evaluated GPVI in two different settings: density estimation and BNNs. In our density estimation experiment, we trained our generator to draw samples from a target distribution, showing that GPVI can accurately fit the posterior from the data. For our BNN experiments, we evaluated on regression, classification, and high dimensional open-category tasks. 
Our experiments show that among all methods compared, GPVI is the only method to excel in both sampling efficiency and asymptotic performance.
We compare GPVI with ParVI methods: SVGD \cite{liu2016stein}, GFSF \cite{liu2019understanding}, and KSD \cite{hu2018stein, grathwohl2020learning}, as well as their corresponding amortized versions.
We also compared with a mean-field VI approach Bayes by Backprop (MF-VI) \cite{blundell2015weight}, as well as deep ensembles \cite{lakshminarayanan2017simple}, and HMC \cite{neal2011mcmc}. 
We emphasize that our aim is not to only maximize the likelihood of generated samples, rather we want to closely approximate the posterior of parameterized functions given the data. 
Thus, predictions w.r.t data unseen during training should reflect the epistemic uncertainty of the model. 

The details of our experimental setup are as follows. In our BNN experiments, we parameterized samples from the target distribution as neural networks with a fixed architecture. 
In the low dimensional regression and classification settings, we drew 100 samples from the approximate posterior for both training and evaluation, allowing us to evaluate the predictive mean and variance. For methods without a sampler e.g. ParVI and deep ensembles, we initialized a 100 member ensemble.
In the high dimensional open-category tasks, we instead used 10 samples due to the larger computation cost. 
For methods that utilize a hypernetwork such as GPVI and amortized ParVI, we used Gaussian input noise $z \sim \mathcal{N}(0, I)$ and varied the hypernetwork architecture depending on the task. 
For ParVI and deep ensembles we randomly initialized 
each member of the ensemble.
In all methods except HMC we used the Adam optimizer \citep{kingma2014adam}. 
A detailed description of network architectures, chosen hyperparameters, and experimental settings is given in the appendix.

\subsection{Density Estimation}
We first evaluate the ability of generative approaches to fit a target distribution from data. We used 2D and 5D zero mean Gaussian distributions with non-diagonal covariances as our target distributions. 
We consider sampler networks with one hidden layer of width 2 and 5, respectively, and input noise with the same dimensionality as the output. The error is defined as the difference between the true variance and the variance of the sampler output distribution $WW^T - \Sigma$, where $W$ is the weights of the sampler. 
As shown in Table~\ref{tab:density_estimation},
while the estimation problem becomes harder with increased dimension,
our approach performs consistently well while amortized ParVI methods suffer more or less from a performance drop and all end up inferior to our approach. We have included more density estimation results in the Appendix.

\begin{table}[]
\centering
\begin{tabular}{|c|c|c|}
\hline
Method  & $\Sigma$ error (2d) $\downarrow$ & $\Sigma$ error (5d) $\downarrow$ \\ \hline
Amortized SVGD & 0.10 $\pm$ .09 & 0.37 $\pm$ .32  \\ \hline
Amortized GFSF & 0.18 $\pm$ .04 & 0.21 $\pm$ .13  \\ \hline
Amortized KSD  & 0.28 $\pm$ .32 & 1.68 $\pm$ .52  \\ \hline
GPVI           & 0.14 $\pm$ .08 & 0.14 $\pm$ .04  \\ \hline
\end{tabular}
\caption{Comparison of generative Particle VI approaches for density estimation of 2d and 5d Gaussian distributions. }
\vskip -0.1in
\label{tab:density_estimation}
\end{table}

\subsection{Bayesian Linear Regression}
We evaluated all methods on Bayesian linear regression to investigate how well each method can fit a unimodal normal distribution over linear function weights. 
The target function is a linear regressor with parameter vector $\vec{\beta}\in \R^d$, i.e., $\vec{y} = \vec{X}\vec{\beta} + \epsilon$, where $\vec{X}\sim\mathcal{N}(0, I_d),\ \epsilon \sim \mathcal{N}(0, I)$, and $\beta^i \sim U(0, 1) + 5$. We chose such linear Gaussian settings where we can explicitly compute the target posterior $p(\vec{\beta}|\vec{X}, \vec{y})\sim \mathcal{N}((\vec{X}^T\vec{X})^{-1}\vec{X}^T\vec{y}, \vec{X}^T\vec{X})$ given observations $(\vec{X}, \vec{y})$, allowing us to numerically evaluate how well each method fits the target distribution. 
Each method was trained to regress $\vec{y}$ from $\vec{X}$. For GPVI and amortized ParVI methods that make use of a hypernetwork, we used a linear generator with input noise $\vec{z} \in \R^d$. In this setting, the output of a generator with bias $\vec{b}$ and weights $W$ follows the distribution $\mathcal{N}(\vec{b}, WW^T)$, which should match the posterior $p(\vec{\beta}|\vec{X},\vec{y})$. 
For ParVI methods and deep ensembles, we initialized an ensemble of linear regressors with $d$ parameters. We computed the mean and covariance of the learned parameters to measure the quality of posterior fit.
For MF-VI we chose a standard normal prior on the weights, and likewise computed the mean and covariance of weight samples to measure posterior fit. We also test a variant of GPVI where we exactly compute the Jacobian inverse instead of using our helper network. We set $d=3$ in our experiments. 
While simple, this quantitative sanity check is crucial before evaluating on more complicated domains where the true posterior is not available. 
It will be evident in later experiments that the ability to closely approximate the true posterior under this simple setup is indicative of performance on higher dimensional tasks.  

\begin{table}[]
\centering
\begin{tabular}{|c|c|c|}
\hline
Method  & $\mu$ error $\downarrow$ & $\Sigma$ error $\downarrow$ \\ \hline
SVGD           & 0.006 $\pm$ .0024 & \textbf{0.125} $\pm$ \textbf{.03}  \\ \hline
GFSF           & 0.003 $\pm$ .0014 & 0.139 $\pm$ .07  \\ \hline
KSD            & 0.009 $\pm$ .0006 & 0.373 $\pm$ .11  \\ \hline

Amortized SVGD & \textbf{0.002} $\pm$ \textbf{.0009} & 0.158 $\pm$ .04  \\ \hline
Amortized GFSF & \textbf{0.002} $\pm$\textbf{ .0003} & 0.209 $\pm$ .05  \\ \hline
Amortized KSD  & 0.004 $\pm$ .0007 & 0.430 $\pm$ .10  \\ \hline

MF-VI    & 0.004 $\pm$ .0035 & 0.303 $\pm$ .04      \\ \hline
Deep Ensemble & 0.004 $\pm$ .0002 & 1.0   $\pm$ 0    \\ \hline
HMC            & 0.009 $\pm$ .0003 & 0.181 $\pm$ .05  \\ \hline
GPVI           & \textbf{0.002} $\pm$\textbf{ .0007} & \textbf{0.128} $\pm$ \textbf{.04}  \\ \hline
GPVI Exact Jac & \textbf{0.002} $\pm$ \textbf{.0004} & \textbf{0.106} $\pm$ \textbf{.07}  \\ \hline 

\end{tabular}
\caption{Bayesian linear regression. Reported error is the $L_2$ norm of the difference between the learned mean and covariance parameters, and the ground truth after 50000 iterations.}
\label{tab:blr}
\vskip -0.1in
\end{table}

Results are shown in table \ref{tab:blr}. 
Our GPVI outperforms amortized ParVI, MF-VI, and surprisingly even HMC. 
GPVI is also very competitive with the best ParVI methods, SVGD and GFSF. We note that GPVI is also competitive with the ``Exact-Jac" variant, where we invert the Jacobian explicitly, instead of using the helper network. This ``Exact-Jac" version of GPVI is not scalable, but it serves as a benchmark to show that our helper network is able to closely approximate the exact solution.  
Finally, as deep ensembles lacks any mechanism for correctly estimating the covariance, it performs poorly as expected. 

\subsection{Multimodal Classification}
The previous experiment possesses an analytically known target distribution. However, most problems of interest involve distributions without closed-form expressions. Therefore, we further tested on a 2-dimensional, 4-class classification problem, where each class consists of samples from one component of a mixture distribution. 
The mixture distribution is defined as $p(x) = \sum_{i=1}^4 \mathcal{N}(\mu_i, 0.3)$, with means $\mu_i \in \{(-2,-2), (-2, 2),$ $(2, -2), (2, 2)\}$. We assigned labels $y_i \in \{1, 2, 3, 4\}$ according to the index of the mixture component the samples were drawn from. 
For this task, samples are weight parameters $\vec\theta$ of two-layer neural networks with 10 hidden units in each layer and ReLU activations, denoted by $f_{\vec{\theta}}: \R^2 \rightarrow \R^4$.

\begin{figure*}[ht!]
\centering
    \subfigure[GPVI]{
        \label{fig:gpvi_4class}
        \includegraphics[width=.30\linewidth]{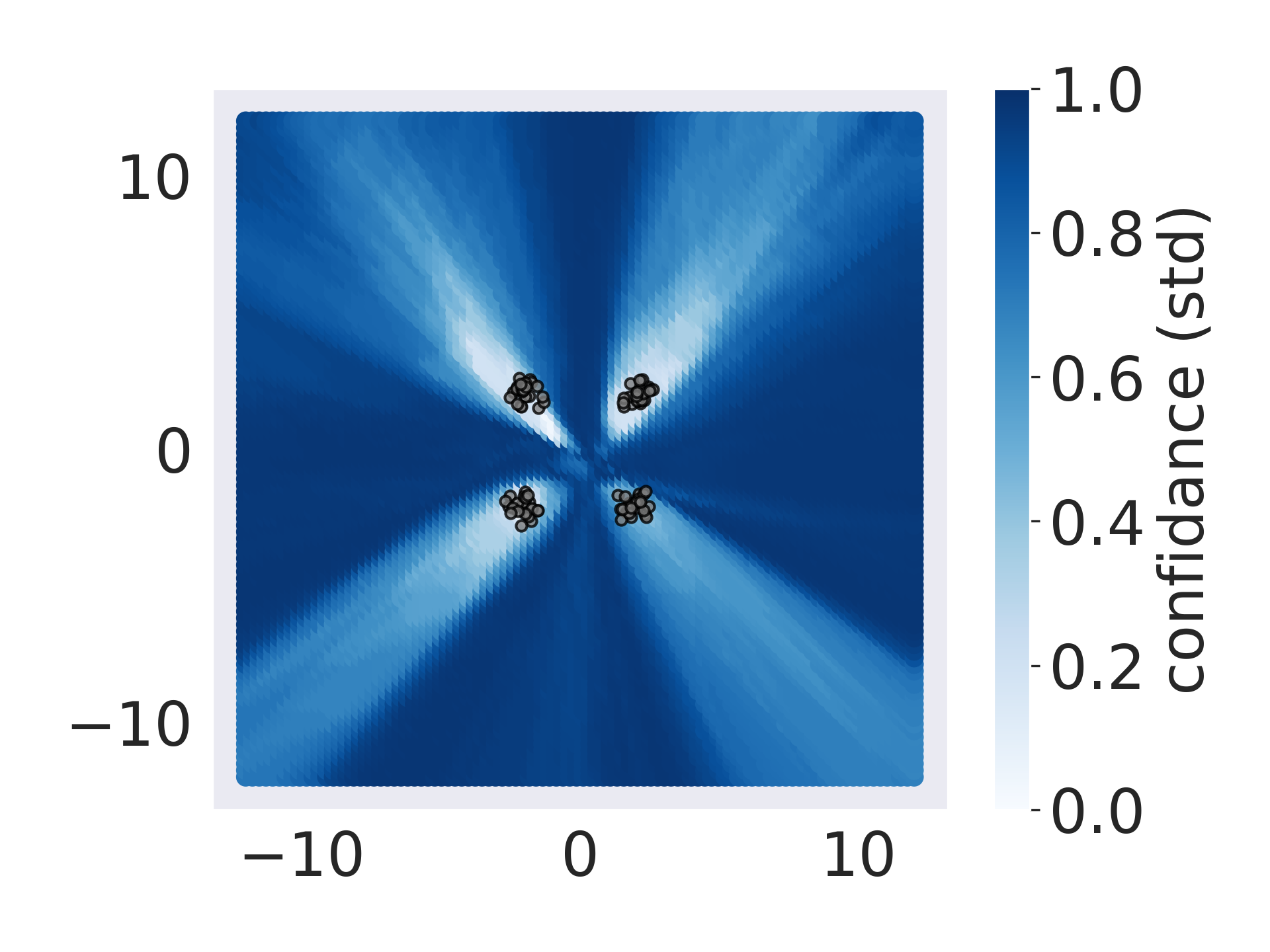}
    }
    \subfigure[SVGD]{
        \label{fig:svgd_4class}
        \includegraphics[width=.30\linewidth]{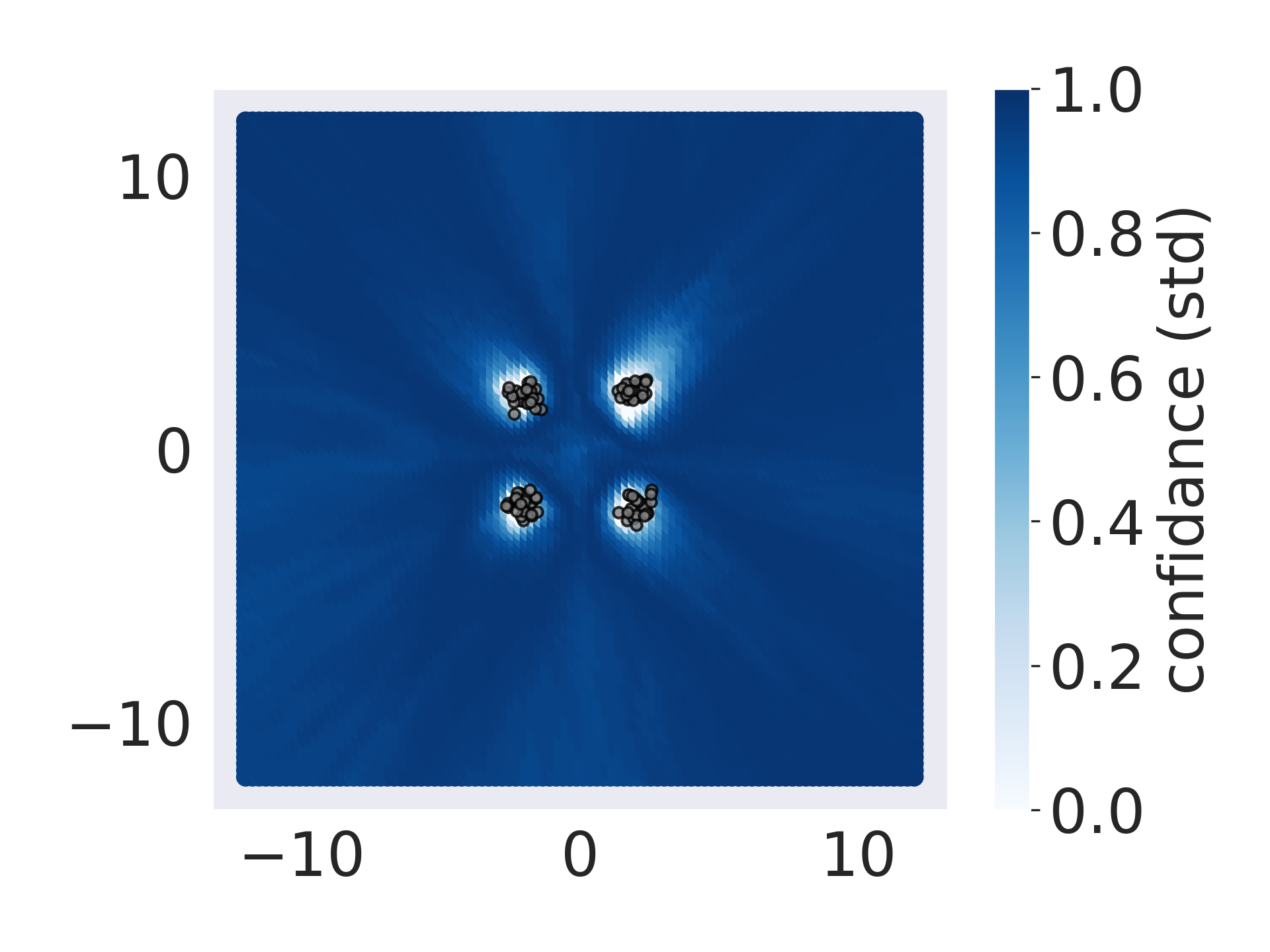}
    }
    \subfigure[HMC]{
        \label{fig:hmc_4class}
        \includegraphics[width=.30\linewidth]{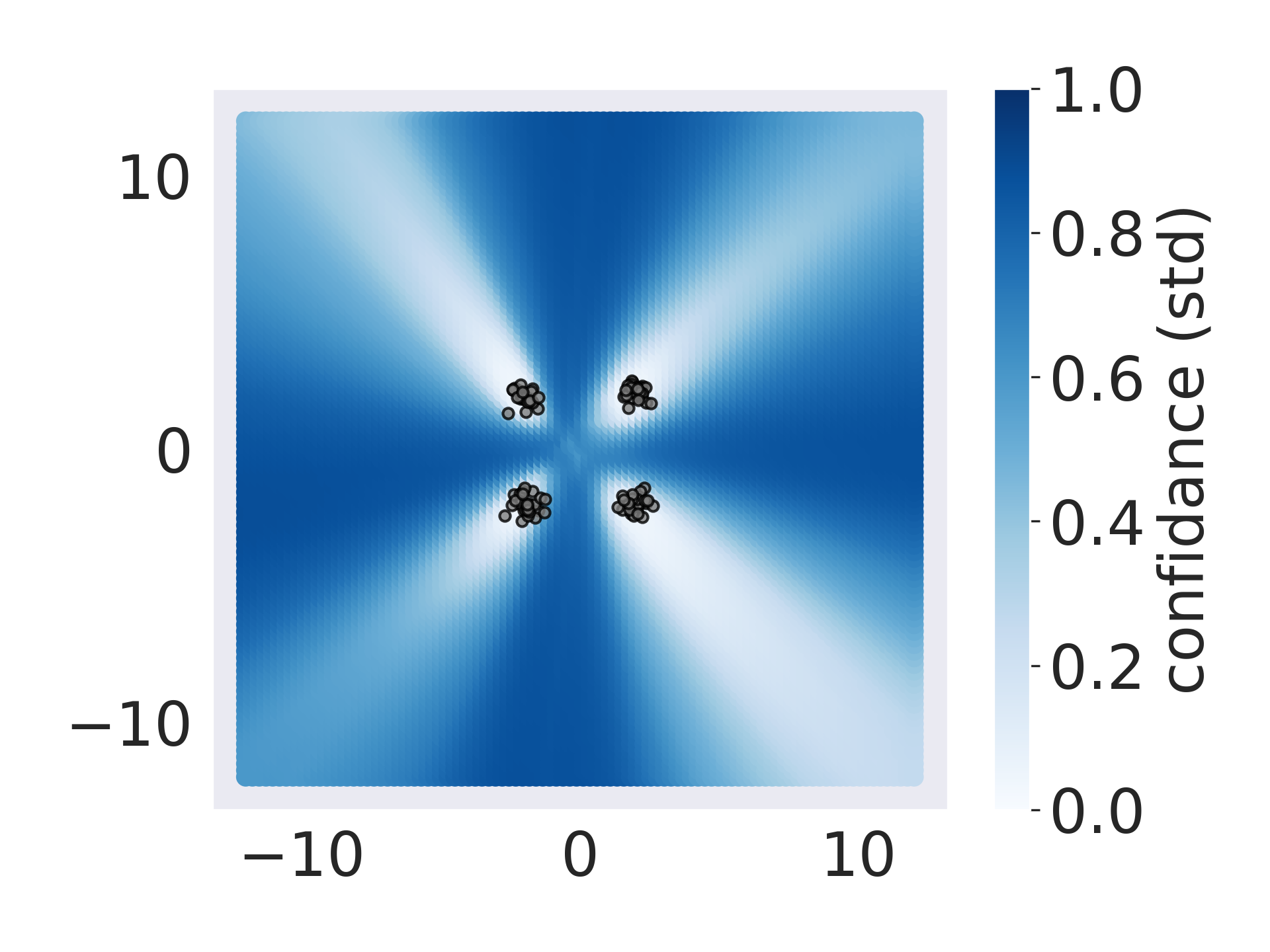}
    }
    \subfigure[Amortized SVGD]{
        \label{fig:asvgd_5class}
        \includegraphics[width=.30\linewidth]{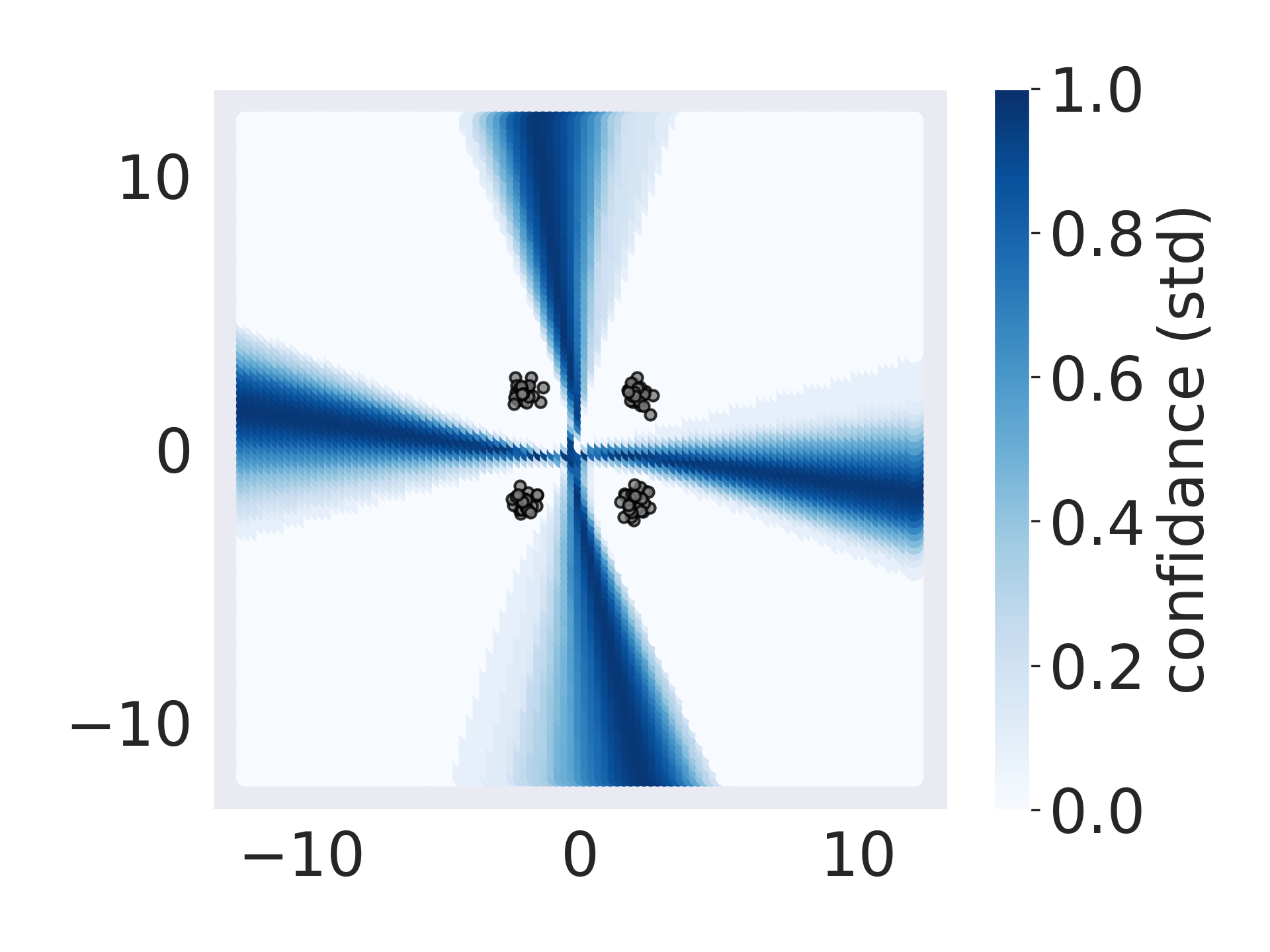}
    }
    \subfigure[MF-VI]{
        \label{fig:mfvi_4class}
        \includegraphics[width=.30\linewidth]{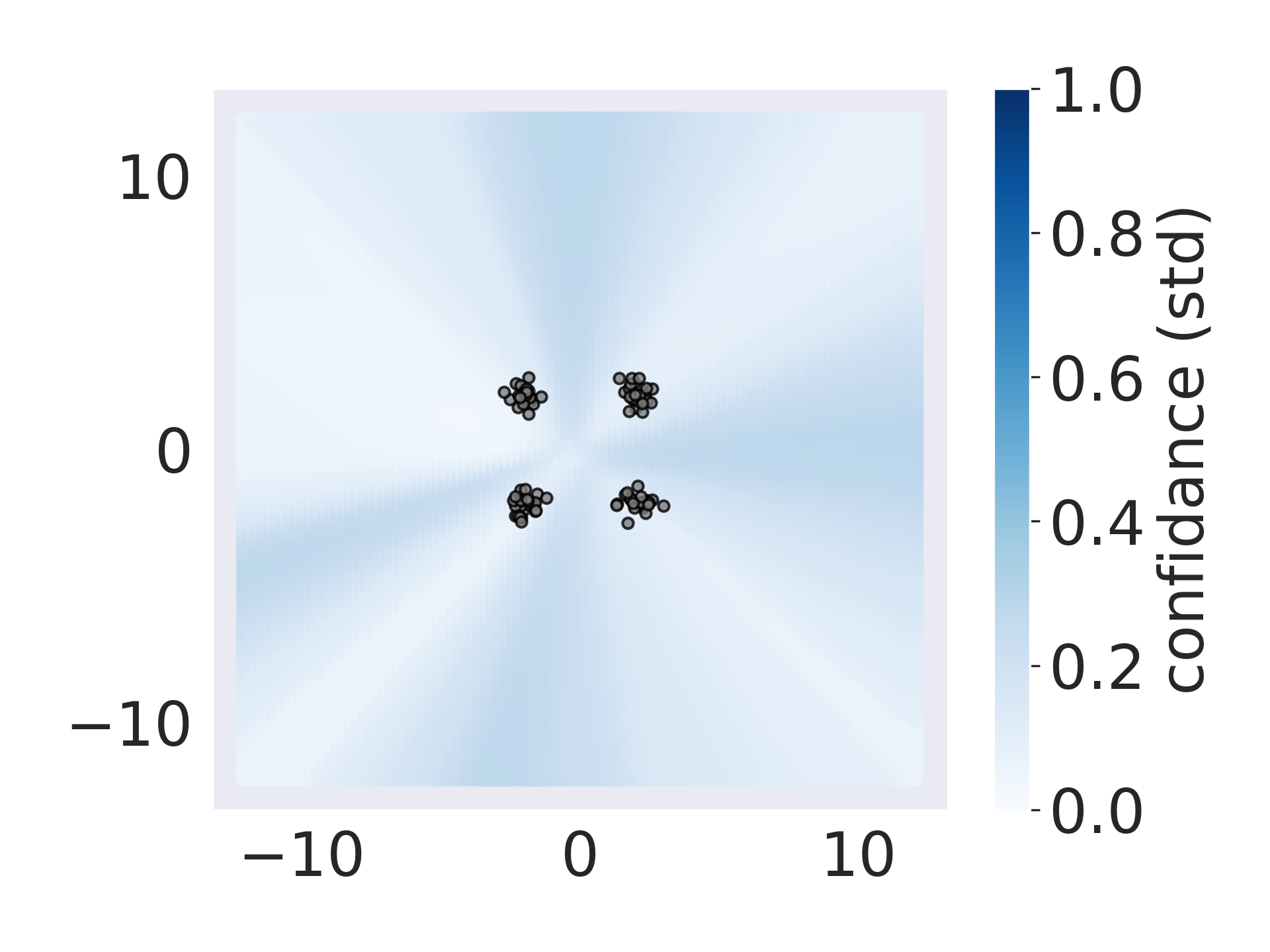}
    }
    \subfigure[Deep Ensemble]{
        \label{fig:mfvi_4class}
        \includegraphics[width=.30\linewidth]{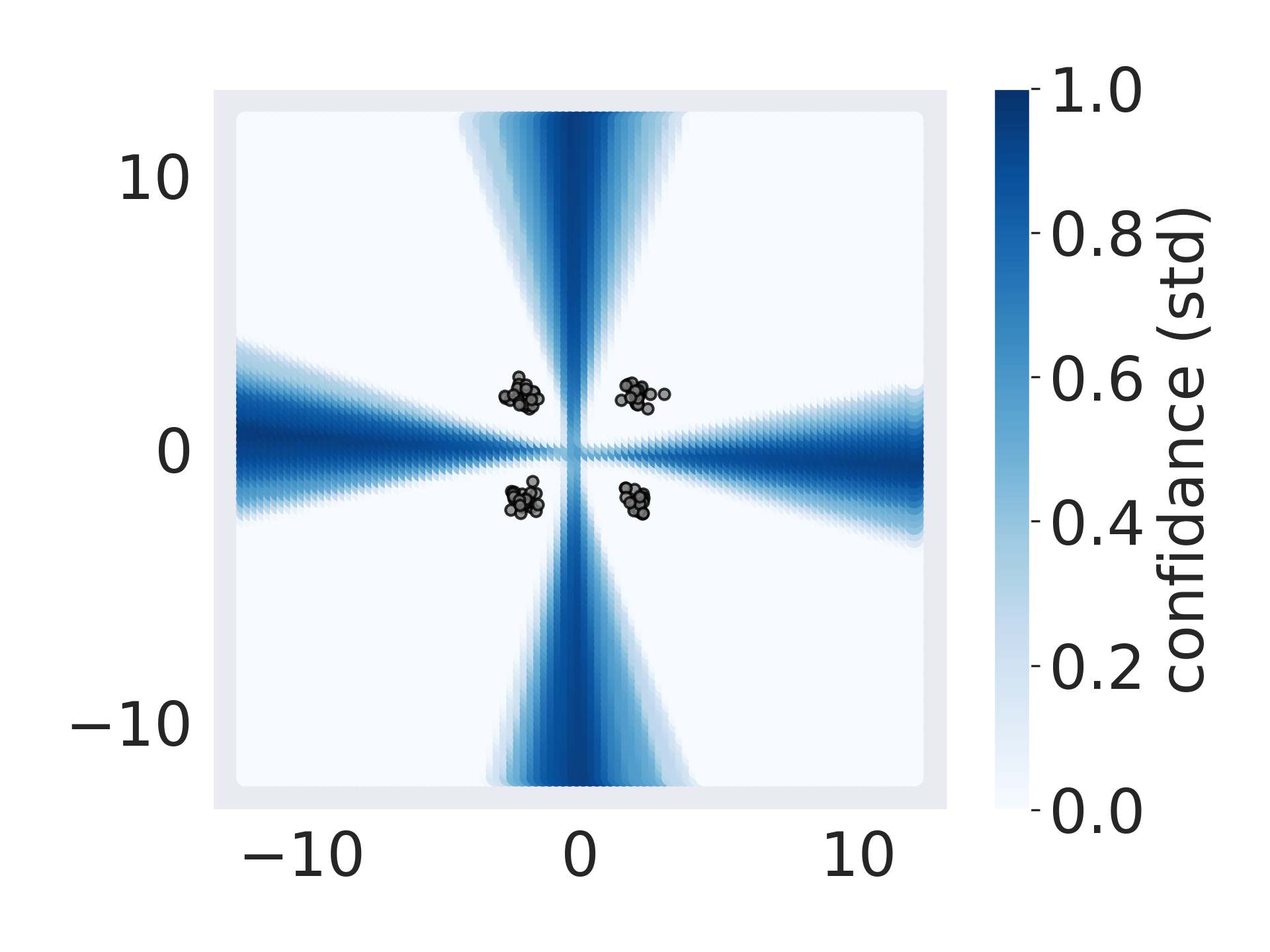}
    }
\vskip -0.1in
\caption{Predictive uncertainty of each method on the 4-class classification task, as measured by the standard deviation between predictions of sampled functions. Regions of high uncertainty are shown as darker, while lighter regions correspond to lower uncertainty. The training data is shown as samples from four unimodal normal distributions. It can be seen that amortized SVGD, MF-VI and deep ensembles significantly underestimate the uncertainty in regions with no training data}   
\label{fig:4class_classification}
\vskip -0.05in
\end{figure*}

To train each method, we drew a total of 100 training points, and 200 testing points from the target distribution. 
To evaluate the posterior predictive distribution, we drew points from a grid spaced from $\{-10, 10\}$, then plot the predictive distribution as measured by the standard deviation in predictions among model samples in figure \ref{fig:4class_classification}.  

In this setting, the true posterior $p(\vec{\theta}|\vec{X}, y)$ is unknown and past work often relies on ``gold-standard" approaches like HMC to serve as the ground truth. In the previous Bayesian linear regression tests, however, we saw that 
HMC was outperformed by GPVI as well as ParVI methods SVGD and GFSF. 
In the current test, intuitively, a ``gold-standard" approach should yield a predictive distribution with high variance (high uncertainty) in regions far from the training data, and low variance (low uncertainty) in regions near each mixture component. 
As shown in figure \ref{fig:4class_classification}, GPVI and SVGD both have higher uncertainty in no-data regions than HMC, while remaining confident on the training data. On the other hand, MF-VI and amortized SVGD both underestimate the uncertainty.  
The performance of MF-VI is as expected, as mean-field VI approaches are known to characteristically underestimate uncertainty. 
Results of all methods as well as more tests on an easier 2-class variant are included in the Appendix.

\subsection{Open Category}
To evaluate the scalability of our approach we turn to large-scale image classification experiments. 
In this setting, it is not possible to exactly measure the accuracy of posterior fits. 
We instead utilize the open-category task to test if our uncertainty estimations can help detect outlier examples. The open category task defines a set of inlier classes that are seen during training, and a set of outlier classes only used for evaluation. 
While in principle, the content of the outlier classes can be arbitrary, we use semantically similar outlier classes by splitting the training dataset into inlier and outlier classes. 
The open-category task is in general more difficult than out-of-distribution experiments where different datasets are used as outlier classes, since distributions of categories from the same dataset may be harder to discriminate.

We evaluated on the MNIST and CIFAR-10 image datasets, following \citet{neal2018open} to split each dataset into 6 inlier classes and 4 outlier classes. 
We performed standard fully supervised training on the 6 inlier classes, and measured uncertainty in the 4 unseen outlier classes.
We evaluated the uncertainty using two widely-used statistics: area under the ROC curve (AUC), and the expected calibration error (ECE). 
The AUC score measures how well a binary classifier discriminates between predictions made on inlier inputs, vs predictions made on outlier inputs. 
A perfect AUC score of 1.0 indicates a perfect discrimination, while a score of 0.5 indicates that the two sets of predictions are indistinguishable. 
ECE partitions predictions into equally sized bins, and computes the $L_1$ difference in expected accuracy and confidence between bins, which represents the calibration error of the bin. 
ECE computes a weighted average of the calibration error of each bin.  
Together AUC and ECE tell us how well a model can detect outlier inputs, as well as how well the model fits the training distribution.

\par\noindent\textbf{MNIST} consists of 70,000 grayscale images of handwritten digits at 28x28 resolution, divided into 60,000 training images and 10,000 testing images. 
We further split the dataset by only using the first six classes for training and testing. 
The remaining four classes are only used to compute the AUC and ECE statistics. 
We chose the LeNet-5 classifier architecture for all models, and trained for 100 epochs.
Due to the larger computational burden, we only consider 10 samples from each method's approximate posterior for both training and evaluation. Note that our approach is capable of generating many more, but it would be computationally costly to train much more for ensembles and particle VI approaches. 
For GPVI and amortized ParVI methods we used a 3 layer MLP hypernetwork with layer widths $[256, 512, 1024]$, ReLU activations, and input noise $\vec{z} \in \R^{256}$. 
We did not test HMC in this setting, as the computational demand is too high. 

Table \ref{tab:mnist-open-category} shows the results. 
All RKHS-based methods (GPVI, ParVI, amortized ParVI) as well as deep ensembles achieve high supervised (``clean``) accuracy with the LeNet architecture. MF-VI underfits slightly, while KSD struggled to achieve competitive accuracy even after 100 epochs.  
All methods achieved an AUC over 0.95, but SVGD, GFSF and GPVI have the highest AUC values, respectively. In terms of calibration, GPVI and RKHS-based ParVI methods are the best calibrated. KSD is the worst calibrated model, and deep ensembles/MF-VI have middling performance. 

\begin{table}[htb]
\centering
\begin{tabular}{|c|c|c|c|}
\hline
Method   & Clean & AUC$\uparrow$   & ECE $\downarrow$ \\ \hline
SVGD           & 99.3 & \textbf{.989} $\pm$ \textbf{.001} & \textbf{.001} $\pm$ \textbf{.0002}  \\ \hline
GFSF           & 99.2 & \textbf{.988} $\pm$ \textbf{.003} & .002  $\pm$ .0003 \\ \hline
KSD            & 97.7 & .964 $\pm$ .005 & .014  $\pm$ .0007 \\ \hline

Amortized SVGD & 99.1 & .958 $\pm$ .015 & \textbf{.002} $\pm$ \textbf{.0007}  \\ \hline
Amortized GFSF & 99.2 & .978 $\pm$ .005 & .004  $\pm$ .0013 \\ \hline
Amortized KSD  & 97.7 & .951 $\pm$ .008 & .017  $\pm$ .0010 \\ \hline

MF-VI     & 98.6 & .951 $\pm$ .008 & .014 $\pm$ .0027  \\ \hline
Deep Ensemble & 99.3 & .972 $\pm$ .002 & .008 $\pm$ .0060  \\ \hline
GPVI           & 99.3 & \textbf{.988} $\pm$ \textbf{.001} & \textbf{.001} $\pm$ \textbf{.0005}  \\ \hline

\end{tabular}
\caption{Results for open-category classification on MNIST. We show the result of standard supervised training (Clean), as well as AUC and ECE statistics computed from training on a subset of classes and testing on the rest of the classes as outliers.}
\label{tab:mnist-open-category}
\end{table}

\par\noindent\textbf{CIFAR-10} consists of 60,000 RGB images depicting 10 object classes at 32x32 resolution, 
divided into 50,000 training images and 10,000 testing images.
We adopted the same 6 inlier / 4 outlier split used in ~\cite{neal2018open} for the open-category setting.
For this task we used 
a CNN with 3 convolutional layers and two linear layers,
which is much smaller than SOTA classifiers for CIFAR-10. 
Though the classification accuracy would suffer, it allows us to clearly evaluate our method without considering interactions with architectural components such as BatchNorm or residual connections. 
We also used 10 samples from each method's approximate posterior for training and evaluation. For GPVI and amortized ParVI methods we used the same hypernetwork architecture as in the MNIST setup. 
Table \ref{tab:cifar10-open-category} shows the results where it can be seen that our GPVI almost match the performance of SVGD and GFSF in terms of AUC while doing a bit better on ECE. 

\begin{table}[]
\centering
\begin{tabular}{|c|c|c|c|}
\hline
Method  & Clean & AUC $\uparrow$   & ECE $\downarrow$ \\ \hline
SVGD            & 80.3  & \textbf{.683} $\pm$ \textbf{.008} & .055 $\pm$ .004 \\ \hline
GFSF            & 80.6  & \textbf{.681} $\pm$ \textbf{.004} & .068 $\pm$ .012 \\ \hline

Amortized SVGD  & 71.12 & .636 $\pm$ .018 & .073 $\pm$ .029 \\ \hline
Amortized GFSF  & 71.09 & .583 $\pm$ .007 & .042 $\pm$ .029 \\ \hline

MF-VI      & 70.0  & .649 $\pm$ .006 & \textbf{.016} $\pm$ \textbf{.002}\\ \hline
Deep Ensemble  & 73.54 & .652 $\pm$ .018 & .033 $\pm$ .011\\ \hline
GPVI            & 76.2  & .677 $\pm$ .008 & \textbf{.018} $\pm$ \textbf{.015}\\ \hline

\end{tabular}
\caption{Open-category classification on CIFAR-10. We show results of standard supervised training (Clean), as well as AUC and ECE of each method trained in the open-category setting. }
\label{tab:cifar10-open-category}
\vskip -0.05in
\end{table}

\subsection{Discussion}
Much of the focus regarding recent work on Bayesian neural networks concerns their performance on open-category and out-of-distribution tasks with high dimensional image datasets. 
Instead, we show that our method closely approximates the target posterior, both in tasks where the posterior is explicitly known, as well as when it is intractable. 
The Bayesian linear regression and density estimation tasks  served as sanity checks. Because the posterior was known explicitly, we could quantitatively test how well each method fit the posterior. For density estimation, GPVI outperforms amortized approaches at matching the target covariance (Table  \ref{tab:density_estimation}). In Bayesian linear regression, while the approximation was close for all methods, there was a clear hierarchy in terms of which types of methods produced the tightest approximation (Table \ref{tab:blr}). 
GPVI and RKHS-based ParVI achieved the best posterior fit overall, and we see in further experiments that quality of fit here is indicative of performance in more difficult tasks. 

The four-class classification problem (Fig. \ref{fig:4class_classification}), while seemingly simple, is particularly difficult for most methods we evaluated since many methods tend to overgeneralize to the corners. 
MF-VI underestimates the uncertainty as expected \citep{yao2019quality, minka2001family, bishop2006pattern}. 
Notably, amortized SVGD also underestimates the uncertainty with a predictive distribution resembling that of a standard ensemble. 
We believe this is due to the compounding approximation error explained in Section \ref{sec:amortized_par_vi}, i.e., naively back-propagating the Stein variational gradient to update the generator is more likely to end up with a non-descent direction compared with back-propagating the exact functional gradient as GPVI.
As shown in Fig.~\ref{fig:4class_classification}, the posterior approximation of GPVI is tighter, with uncertainty that better matches the data distribution. 
SVGD performs the best in this task, with high uncertainty everywhere except in regions near observed data. Surprisingly, GPVI and SVGD outperform HMC here, with clearly higher uncertainty near the corners of the sample space. Note that with the functional approximation by neural networks, it is hard to determine if the true posterior exactly matches the intuitively ``ideal" uncertainty plot where low variance only shows up around each mixture component.
On the other hand, HMC is guaranteed to converge to the true posterior over time \cite{durmus2017convergence}. Given a fixed computational budget, it is possible that GPVI or SVGD could achieve better performance. 
GPVI also has the extra benefit of the ability to draw additional samples, which is not possible with ParVI or HMC. 

Our final experimental setting, open-category, reveals that GPVI consistently has a higher AUC than all other scalable sampling approaches.
On MNIST, GPVI is among the overall top performers, together with two ParVI methods SVGD and GFSF. On CIFAR-10, GPVI is among the top two performers in ECE and performs only slightly behind the two best ParVI methods in AUC. Most importantly, GPVI outperforms with clear margins all amortized ParVI methods on both datasets under all metrics.
This is consistent with all other qualitative and quantitative experiments we have conducted, which again shows the advantage of GPVI over existing generative ParVI methods.

We found that KSD completely failed in CIFAR-10, achieving slightly better than random accuracy. We believe this is due to the way that KSD performs the particle update. Where SVGD has a closed form expression for the particle transportation map, KSD parameterizes it with a critic network that has the input-output dimensionality of the particle parameters. Training this critic naturally becomes difficult when applying to neural network functions.

In the open-category setting, there is added time complexity for GPVI relative to amortized methods due to training our helper network. Compared to amortized SVGD, the extra time complexity of training GPVI is a
constant factor less than 1. In our MNIST experiments, this additive factor is 0.28; in CIFAR-10 experiments, this factor is 0.78. 
The difference in MNIST and CIFAR-10 is due to that the output dimension of the helper network scales with the output dimension of the generator. We believe this is acceptable given the
performance gain and the possible real (offline training) applications of GPVI.

\section{Conclusion}
We have presented a new method that fuses the best aspects of parametric VI with non-parametric ParVI. GPVI has asymptotic convergence on par with ParVI. Additionally, GPVI can efficiently draw samples from the posterior.
We also presented a method for efficiently estimating the product between the inverse of the Jacobian of a deep network, and a gradient vector. Our experiments showed that GPVI performs on par with ParVI,
and outperforms amortized ParVI and other competing methods
in Bayesian linear regression, a classification task, as well as open-category tasks on MNIST and CIFAR-10. In the future we want to explore the efficacy of our method applied to large scale tasks like image generation.

\subsection*{Acknowledgements}
Neale Ratzlaff and Li Fuxin were partially supported by the Defense Advanced Research Projects Agency (DARPA) under Contract No. HR001120C0011 and HR001120C0022. Any opinions, findings and conclusions or recommendations expressed in this material are those of the author(s) and do not necessarily reflect the views of DARPA.
\bibliography{references}
\bibliographystyle{icml2021}

\appendix
\onecolumn
\section*{Appendix for Generative Particle Variational Inference via Estimation of Functional Gradients}
\vspace{.5in}

\section{Proofs}
The summation convention is used on all repeated indices (e.g. $a^i b^i := \sum_i a^i b^i$) in the following proofs.

\subsection{Proof of Theorem 3.1}
To compute the gradient of $\mathcal{J}(\vec{f}) = \mathbf{E}_{\vec{z}}\left[-\log p(\vec{f}(\vec{z})) 
 + \log\frac{p_{\vec{z}}(\vec{z})}
 {\left|\det\left(\frac{\partial\vec{f}}{\partial\vec{z}}\right)\right|} \right],$
for any $\vec{\phi}\in T_{\vec{f}}\mathcal{H}$,
\begin{eqnarray}
\label{eq:func_diff}  
d\mathcal{J}_{\vec{f}}(\vec{\phi} )
&=& \frac{d}{dt}\bigg|_{t=0}\mathcal{J}(\vec{f}+t\vec{\phi}) \nonumber\\
&\stackrel{\numcircledtikz{1}}{=}& \E_{\vec{z}}\left[ \frac{d}{dt}\bigg|_{t=0} 
    \left( -\log p((\vec{f}+t\vec{\phi})(\vec{z})) \right) \right] 
    - \E_{\vec{z}}\left[ \frac{d}{dt}\bigg|_{t=0}\log \left| \det\left(
        \frac{\partial(\vec{f}+t\vec{\phi})}{\partial\vec{z}}\right)\right| \right] \nonumber\\
&=& \E_{\vec{z}}\left[ -\nabla_{x^i}\log p(\vec{x})\bigg|_{\vec{x}=\vec{f}(\vec{z})}
    \frac{d}{dt}\bigg|_{t=0}(f^i+t\phi^i)(\vec{z}) \right] 
    - \E_{\vec{z}}\left[ 
    \Tr\left( \left(\frac{ \partial\vec{f} }{ \partial\vec{z} } \right)^{-1}
      \frac{d}{dt}\bigg|_{t=0}\frac{\partial (\vec{f}+t\vec{\phi})}{\partial\vec{z}} \right) 
    \right] \nonumber\\
&=& \E_{\vec{z}}\left[ -\nabla_{x^i}\log p(\vec{x})\bigg|_{\vec{x}=\vec{f}(\vec{z})}
    \phi^i(\vec{z}) \right] 
    - \E_{\vec{z}}\left[ 
    \left( \left(\frac{ \partial\vec{f} }{ \partial\vec{z} } \right)^{-1} \right)^j_i
      \frac{\partial\phi^i}{\partial z^j}  
    \right] \nonumber\\
&\stackrel{\numcircledtikz{2}}{=}& 
    \E_{\vec{z}}\left[ -\nabla_{x^i}\log p(\vec{x})\bigg|_{\vec{x}=\vec{f}(\vec{z})}
    \langle k(\vec{z}, \cdot),\ \phi^i(\vec{z}) \rangle_{\mathcal{H}} \right]
    - \E_{\vec{z}}\left[ \left\langle 
    \left( \left(\frac{ \partial\vec{f} }{ \partial\vec{z} } \right)^{-1} \right)^j_i
    \nabla_{z^j}k(\vec{z}, \cdot),\ \phi^i(\vec{z}) \right\rangle_{\mathcal{H}} 
    \right]  \nonumber\\
&=& \Bigg\langle \E_{\vec{z}}\Bigg[ 
    -\nabla_{x^i}\log p(\vec{x})\bigg|_{\vec{x}=\vec{f}(\vec{z})} k(\vec{z}, \cdot) 
    - \left( \left(\frac{ \partial\vec{f} }{ \partial\vec{z} } \right)^{-1} \right)^j_i
    \nabla_{z^j}k(\vec{z}, \cdot) \Bigg], 
    \phi^i \Bigg\rangle_{\mathcal{H}},
\end{eqnarray}
the following identities are used in step $\numcircledtikz{1}$ and $\numcircledtikz{2}$,
$$d\log|\det A| = \Tr(A^{-1}dA),$$
$$\phi^i(\vec{z})=\langle k(\vec{z},\cdot), \phi^i(\vec{z})\rangle_H,$$
$$\nabla_{z^j}\phi^i(\vec{z}) = \langle\nabla_{z^j}k(\vec{z},\cdot), \phi^i(\vec{z})\rangle_H.$$
By definition of the gradient, 
\begin{equation}
\label{eq:func_grad}
\nabla_{\vec{f}}\mathcal{J}(\vec{f})(\vec{z})
=\mathbf{E}_{\vec{z}'}\left[ -\nabla_{\vec{x}}\log p(\vec{x})\bigg|_{\vec{x}=\vec{f}(\vec{z}')}
    k(\vec{z}', \vec{z})
    - \left( \left(\frac{ \partial\vec{f} }{ \partial\vec{z}' } \right)^{-1} \right)
    \nabla_{\vec{z}'}k(\vec{z}', \vec{z})
    \right].
\end{equation}

\subsection{Proof of Lemma 3.2}
To compute the gradient of 
$\mathcal{J}(\vec\phi) = \mathbf{E}_{\vec{z}}\left[-\log p(\vec\phi\left(\vec{f}(\vec{z}))\right) 
 + \log\frac{p_{\vec{z}}(\vec{z})}
 {\left| \det\left(\frac{\partial(\vec\phi\circ\vec{f})}{\partial\vec{z}}\right) \right| }\right]$ 
at $\vec\phi=\text{id}$,
for any $\vec{v}\in T_{\vec\phi}\mathcal{H}$,

\begin{eqnarray}
\label{eq:func_diff}  
d\mathcal{J}_{\vec\phi}(\vec{v} )
&=& \frac{d}{dt}\bigg|_{t=0}\mathcal{J}(\vec\phi+t\vec{v})\bigg|_{\vec\phi=\text{id}} \nonumber\\
&=& \E_{\vec{z}}\left[ \frac{d}{dt}\bigg|_{t=0} 
    \left( -\log p((\vec{f}+t\vec{v}\circ\vec{f})(\vec{z})) \right) \right] 
    - \E_{\vec{z}}\left[ \frac{d}{dt}\bigg|_{t=0}\log \left| \det\left(
        \frac{\partial(\vec{f}+t\vec{v}\circ\vec{f})}{\partial\vec{z}}\right)\right| \right] \nonumber\\
&=& \E_{\vec{z}}\left[ -\nabla_{x^i}\log p(\vec{x})\bigg|_{\vec{x}=\vec{f}(\vec{z})}
    \frac{d}{dt}\bigg|_{t=0}(f^i+t(\vec{v}\circ\vec{f})^i(\vec{z}) \right] 
    - \E_{\vec{z}}\left[ 
    \Tr\left( \left(\frac{ \partial\vec{f} }{ \partial\vec{z} } \right)^{-1}
      \frac{d}{dt}\bigg|_{t=0}\frac{\partial (\vec{f}+t\vec{v}\circ\vec{f})}{\partial\vec{z}} \right) 
    \right] \nonumber\\
&=& \E_{\vec{z}}\left[ -\nabla_{x^i}\log p(\vec{x})\bigg|_{\vec{x}=\vec{f}(\vec{z})}
    v^i(\vec{f}(\vec{z})) \right] 
    - \E_{\vec{z}}\left[ 
    \left( \left(\frac{ \partial\vec{f} }{ \partial\vec{z} } \right)^{-1} \right)^j_i
      \frac{\partial f^k}{\partial z^j}\frac{\partial v^i}{\partial x^k}
    \right] \nonumber\\
&=& \E_{\vec{z}}\left[ -\nabla_{x^i}\log p(\vec{x})\bigg|_{\vec{x}=\vec{f}(\vec{z})}
    \langle k(\vec{f}(\vec{z}), \cdot),\ v^i(\vec{f}(\vec{z})) \rangle_{\mathcal{H}} \right]
    - \E_{\vec{z}}\left[ \left\langle \nabla_{x^i}k(\vec{x}, \cdot)\bigg|_{\vec{x}=\vec{f}(\vec{z})},\ 
      v^i(\vec{f}(\vec{z})) \right\rangle_{\mathcal{H}} \right]  \nonumber\\
&=& \bigg\langle \E_{\vec{x}=\vec{f}(\vec{z})}\left[ 
    - \nabla_{x^i}\log p(\vec{x}) k(\vec{x}, \cdot) 
    - \nabla_{x^i}k(\vec{x}, \cdot) \right], v^i \bigg\rangle_{\mathcal{H}}.
\end{eqnarray}
By definition of the gradient, 
\begin{equation}
\label{eq:svgd_func_grad}
    \nabla_{\vec\phi}\mathcal{J}(\vec\phi)(\vec{x})\Big|_{\vec\phi=\text{id}}
    = \mathbf{E}_{\vec{x}'}\left[ -\nabla_{\vec{x}'}\log p(\vec{x}') k(\vec{x}', \vec{x}) 
    - \nabla_{\vec{x}'}k(\vec{x}', \vec{x})
    \right],
\end{equation}
where $\vec{x}=\vec{f}(\vec{z})$ and $\vec{x}'=\vec{f}(\vec{z}')$.

\subsection{Directly computing the gradient $\mathcal{J}(\vec\theta)$}
As a completion of the discussion, we also derive the gradient of 
$\mathcal{J}(\vec\theta) = \mathbf{E}_{\vec{z}}\left[-\log p(\vec{f}_{\vec\theta}(\vec{z})) 
    + \log\frac{p_{\vec{z}}(\vec{z})}
        {\left|\det\left(\frac{\partial\vec{f}_{\vec\theta}}{\partial\vec{z}}\right)\right|}\right]$ w.r.t.
$\vec\theta$, for any $\vec{v}\in T_{\vec\theta}W$,

\begin{eqnarray}
\label{eq:param_grad}  
d\mathcal{J}_{\vec{\theta}}(\vec{v} )
&=& \frac{d}{dt}\bigg|_{t=0}\mathcal{J}(\vec{\theta}+t\vec{v}) \nonumber \\
&=& \E_{\vec{z}}\left[ \frac{d}{dt}\bigg|_{t=0} 
    \left( -\log p(\vec{f}_{\vec{\theta}+t\vec{v}}(\vec{z})) \right) \right]
    - \E_{\vec{z}}\left[ \frac{d}{dt}\bigg|_{t=0}\log\det\left(
        \frac{\partial\vec{f}_{\vec{\theta}+t\vec{v}}}{\partial\vec{z}}\right) \right] \nonumber \\
&=& \E_{\vec{z}}\left[ -\nabla_{x^j}\log p(\vec{f}_{\vec{\theta}}(\vec{z}))
    \frac{d}{dt}\bigg|_{t=0} f^j_{\vec{\theta}+t\vec{v}}(\vec{z}) \right]
    - \E_{\vec{z}}\left[ 
    \Tr\left( \left(\frac{ \partial\vec{f}_{\vec{\theta}} }{ \partial\vec{z} } \right)^{-1}
      \frac{d}{dt}\bigg|_{t=0}\frac{\partial\vec{f}_{\vec{\theta}+t\vec{v}}}{\partial\vec{z}} \right) 
    \right] \nonumber \\
&=& \E_{\vec{z}}\left[ -\nabla_{x^j}\log p(\vec{f}_{\vec{\theta}}(\vec{z}))
    \frac{\partial f^j}{\partial\theta^i}v^i i\right]
    - \E_{\vec{z}}\left[ 
    \Tr\left( \left(\frac{ \partial\vec{f}_{\vec{\theta}} }{ \partial\vec{z} } \right)^{-1}
      \frac{\partial^2\vec{f}_{\vec{\theta}}}{\partial\vec{z}\partial\vec{\theta}}\vec{v} \right) 
    \right] \nonumber \\
&=& \Bigg\langle \E_{\vec{z}}\Bigg[ -\nabla_{x^j}\log p(\vec{f}_{\vec{\theta}}(\vec{z}))
        \frac{\partial f^j}{\partial\theta^i}
        - \left(\left( \frac{\partial\vec{f}_{\vec{\theta}}}{\partial\vec{z}}\right)^{-1} \right)^j_k
          \left( \frac{\partial^2 f^j_{\vec{\theta}}}{\partial z^k\partial\theta^i} \right)
        \Bigg], v^i
    \Bigg\rangle_E,
\end{eqnarray}
where $\langle\cdot,\cdot\rangle_E$ denotes the Euclidean inner product.
Therefore, 
\begin{equation*}
\nabla_{\theta^i}(\mathcal{J}(\vec\theta))
= \mathbf{E}_{\vec{z}}\left[-\frac{\partial f^j}{\partial\theta^i}
    \nabla_{x^j}\log p(\vec{f}_{\vec{\theta}}(\vec{z}))
    - \left(\left( \frac{\partial\vec{f}_{\vec{\theta}}}{\partial\vec{z}}\right)^{-1} \right)^j_k
      \left( \frac{\partial^2 f^j_{\vec{\theta}}}{\partial z^k\partial\theta^i} \right) \right]. 
\end{equation*}
Computation of the second term in the expectation involves second derivatives of $\vec{f}_{\vec\theta}$, which is not amenable in large-scale problems. As a result, we avoid this direct approach and choose to pullback the functional gradient~\eqref{eq:func_grad} instead.

\subsection{Proof of Injectivity of $\vec{f}$}
We parameterize $\vec{f}$ as follows,
\begin{equation}
\label{eq:generator}
    \vec{f}(\vec{z}) = \vec{g}\left(\vec{z}^{(:k)}\right) + \lambda \vec{z}, \quad \forall\vec{z}\in\R^d,
\end{equation}
where $\vec{z}^{(:k)}\in\R^k$ denotes the vector consisting of the first $k(\ll d)$ components of $\vec{z}$,
and $\vec{g}: \R^k\to\R^d$ is a much slimmer neural network. For any experimental setting, our parameterization of $\vec{g}$ uses a dimension $k$ less than $30\%$ the size of $d$. For our high dimensional open-category experiments where $d > 60000$, we use a $k$ of less than $2\%$ of $d$. 
The Jacobian is,
\begin{equation}
\label{eq:jac_gen}
\left[\frac{\partial\vec{f}}{\partial\vec{z}}\right]_{d\times d} =     
    \left[ \left[\frac{\partial\vec{g}}{\partial\vec{z}^{(:k)}}\right]_{d\times k} 
        \bigg| \vec{0}_{d\times (d-k)} \right]_{d\times d}
    + \lambda \vec{I}_d.
\end{equation}

We now show that our practical choice of $\vec{f}$ is indeed injective.

From~\eqref{eq:generator}, for sufficiently large $\lambda$, the Jacobian
$J_{\vec{f}}$ given by~\eqref{eq:jac_gen} is positive definite. 
Since the domain of $\vec{z}$ is convex, it is straightforward to show $\vec{f}$ is injective,

For any two different points $\vec{z}_1, \vec{z}_2\in\mathcal{Z}$, consider the line segment
$\vec{z}_1 + t(\vec{z}_2 - \vec{z}_1), t\in [0, 1]$, 
which lies in $\mathcal{Z}$ given the convexity of $\mathcal{Z}$. From the Fundamental Theorem of calculus,
\begin{eqnarray*}
(\vec{z}_2 - \vec{z}_1)^T(\vec{f}\left(\vec{z}_2) - \vec{f}(\vec{z}_1)\right)
&=& (\vec{z}_2 - \vec{z}_1)^T 
\left( \int_0^1 J_{\vec{f}}(\vec{z}_1 + t(\vec{z}_2 - \vec{z}_1))dt \right)\cdot (\vec{z}_2 - \vec{z}_1) \\
&=& \int_0^1 (\vec{z}_2 - \vec{z}_1)^T J_{\vec{f}}(\vec{z}_1 + t(\vec{z}_2 - \vec{z}_1)) (\vec{z}_2 - \vec{z}_1) dt > 0
\end{eqnarray*}
the last inequality is due to the positive definiteness of $J_{\vec{f}}$ on $\mathcal{Z}$.
Therefore, $\vec{f}(\vec{z}_1)\neq\vec{f}(\vec{z}_2)$, $\vec{f}$ is injective.

\section{Method Details}
We provide architectures and computational details of our generator and helper networks.

\subsection{Helper Network}
The helper network $\vec{h}_{\eta}$ takes $\vec{z}'$ and 
$\nabla_{\vec{z}'}k(\vec{z}',\vec{z})$ as inputs and outputs 
$\left(\frac{\partial\vec{f}(\vec{z}')}{\partial\vec{z}'}\right)^{-1}
\nabla_{\vec{z}'}k(\vec{z}',\vec{z})$.
Given the specific parameterization of $\vec{f}$ defined by~\eqref{eq:generator}, 
the Jacobian~\eqref{eq:jac_gen} only depends on $\vec{z}^{'(:k)}$,
the first $k$ components of $\vec{z}'$. The helper network applies a fully connected layer with ReLU 
activation to the input $\vec{z}^{'(:k)}$ and $\nabla_{\vec{z}'}k(\vec{z}',\vec{z})$ respectively, 
then the outputs from these two branches are concatenated and sent to a three-layer fully connected network. We optimize the helper network with the Adam optimizer \citet{kingma2014adam} and a learning rate of $1e-4$. We found that regularization techniques like batchnorm and weight decay hurt performance, presumably because we are optimizing for an exact solution and thus do not want a smoothed estimate. While our helper network makes computation of the Jacobian inverse tractable, it nonetheless adds the complexity of an extra optimization problem per training step of $\vec{f}$. To mitigate the increased training time we update $\vec{h}_\eta$ just once per training step of $\vec{f}$; taking one step also keeps the network from over-committing to a single input. 

\subsection{Kernel Selection}
We follow \citet{liu2016stein} and use an RBF kernel $k(x, x') = \exp{(\frac{1}{h}||x - x'||^2_2)}$ for all kernel-based methods including GPVI, SVGD, GFSF, and their amortized variants. The bandwidth $h$ is computed using the median method also proposed by \citet{liu2016stein}: $h = med^2/\log n$, where $med$ is the median of pairwise distances between samples $\{x_i\}_{i=1}^n$. 

\section{Comparing with an Alternative Solver for $\left(\frac{\partial\vec{f}}{\partial\vec{z}'}\right)^{-1}
\nabla_{\vec{z}'}k(\vec{z}',\vec{z})$}
\begin{figure*}[ht!]
\centering
    \subfigure[Mean Error]{
        \label{fig:solver_mean_err}
        \includegraphics[width=.45\linewidth]{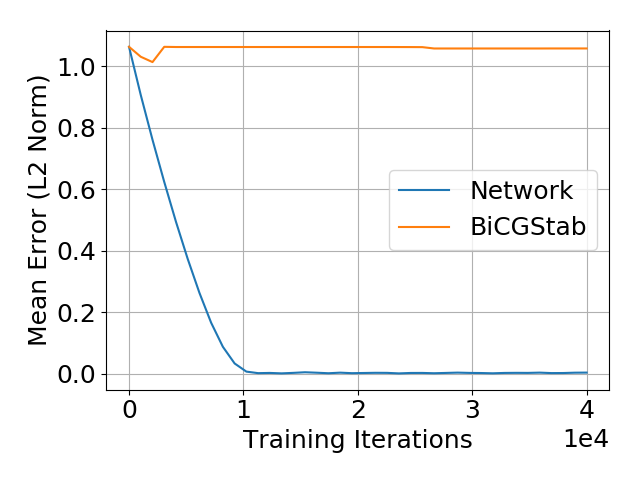}
    }
    \subfigure[Cov Error]{
        \label{fig:solver_cov_err}
        \includegraphics[width=.45\linewidth]{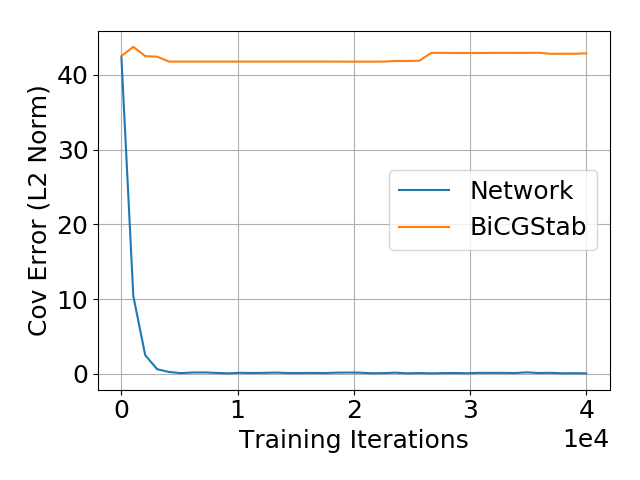}
    }
\vskip -0.1in
\caption{Comparing our helper network with BiCGSTAB in the Bayesian linear regression setting.}   
\label{fig:solver_compare}
\vskip -0.05in
\end{figure*}

Here we justify the use of our helper network $\vec{h}_\eta$, by comparing its performance to a more traditional linear solver: the stabilized biconjugate gradient method (BiCGSTAB) \citep{saad2003iterative}. BiCGSTAB is a well-known iterative algorithm for solving systems of the form $Ax = b$. It is similar to the conjugate gradient method, but does not require $A$ to be self-adjoint, giving BiCGSTAB wider applicability. In figure \ref{fig:solver_compare}, we compare our helper network ("Network") against BiCGSTAB in the Bayesian linear regression setting from section 4.2 in the main text. 
For BiCGSTAB, we solve $B*B$ systems of the form $JJ^{-1} \nabla_{\vec{z}'} k(\vec{z}', \vec{z}) = \nabla_{\vec{z}'} k(\vec{z}', \vec{z})$ for each iteration of training our generator, where $J = \frac{\partial \vec{f}}{\partial \vec{z}'}$ and $B*B$ is the effective batch size of $\nabla_{\vec{z}'}k(\vec{z}', \vec{z})$. 
Due to the greatly increased training time from solving $B*B$ independent problems per training iteration, we only run BiCGSTAB for one step, and warm start from the previous solution at each new generator training iteration. In addition to increased training time, BiCGSTAB suffers from instability due to the constantly changing $\frac{\partial\vec{f}(\vec{z}')}{\partial\vec{z}'}$ as well as $\nabla_{\vec{z'}}k(\vec{z'}, \vec{z})$, due to the resampling of $\vec{z}$ and $\vec{z'}$ at each step. As seen in figure \ref{fig:solver_compare}, when using BiCGSTAB, GPVI is unable to fit the target distribution, while using our helper network ("Network") we are able to efficiently minimize the mean and covariance error. 

We also considered using a normalizing flow as a replacement for our generator network, as normalizing flows are invertible with lower triangular Jacobians. Unfortunately, the efficiency of normalizing flows comes from setting the input-output dimensionality to be equal to force the Jacobian to be square. When using our method for BNNs, as explained in Section B, we cannot afford to store a generator with equal input-output dimensionality, making normalizing flows an inefficient choice at best for our own method. Our helper network stands as an efficient, novel solution for explicitly estimating the Jacobian inverse vector product when needed.  
Nevertheless, we compare GPVI with normalizing flow method for density estimation in the next section.

\section{Additional Results}
We show additional results on density estimation and classification. 

\subsection{Density Estimation}
Here we provide additional results in the density estimation setting. 

\subsubsection*{Comparison with Explicit Jacobian}
Here we perform the experiments from section 4.1 of the main paper, and compare GPVI with a variant where the Jacobian and its inverse are explicitly computed. The inverse is computed with the PyTorch \citep{paszke2019pytorch} function \texttt{torch.inverse}, which uses LAPACK routines \texttt{getrf} and \texttt{getri}. We can see in table~\ref{tab:density_estimation_exact_jac} that GPVI is competitive with the ``Exact-Jac" variant in the 2D setting, and even performs better in the 5D setting. This is unexpected, as explicitly computing the Jacobian and its inverse should be the correct way to compute the functional gradient. We hypothesize that the Jacobian of our generator network is ill-conditioned. In this case any inversion algorithm is more prone to large numerical error. Because our helper network is updated once per training iteration, we may avoid the large gradients that come from numerical error. The increased performance of GPVI over the ``Exact-Jac" variant in the 5D setting may be due to this smoother training.   

\begin{table}[]
\centering
\begin{tabular}{|c|c|c|}
\hline
Method  & $\Sigma$ error (2d) $\downarrow$ & $\Sigma$ error (5d) $\downarrow$ \\ \hline
Amortized SVGD & 0.10 $\pm$ .09 & 0.37 $\pm$ .32  \\ \hline
GPVI           & 0.14 $\pm$ .08 & 0.14 $\pm$ .04  \\ \hline
GPVI Exact Jac & 0.15 $\pm$ .09 & 0.49 $\pm$ .17  \\ \hline
\end{tabular}
\caption{Comparison of generative Particle VI approaches for density estimation of 2d and 5d Gaussian distributions. }
\vskip -0.1in
\label{tab:density_estimation_exact_jac}
\end{table}

\subsubsection*{Energy Potentials}
\begin{figure*}[ht!]
\centering
    \subfigure[Twin Moons Target]{
        \label{fig:twin-moons-target}
        \includegraphics[width=.23\linewidth]{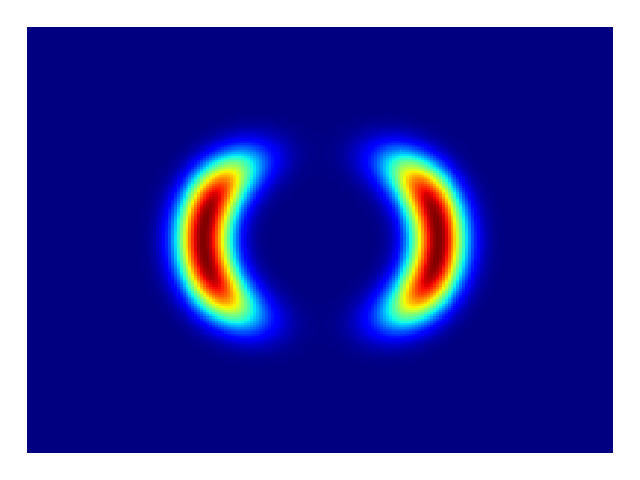}
    }
    \subfigure[GPVI]{
        \label{fig:moons-gpvi}
        \includegraphics[width=.23\linewidth]{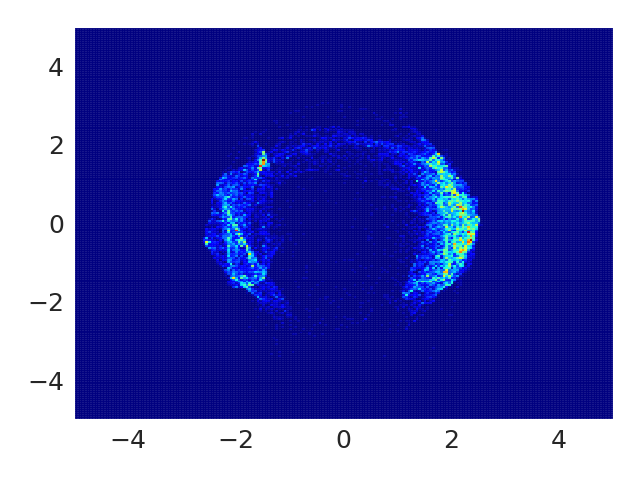}
    }
    \subfigure[Amortized SVGD]{
        \label{fig:moons-asvgd}
        \includegraphics[width=.23\linewidth]{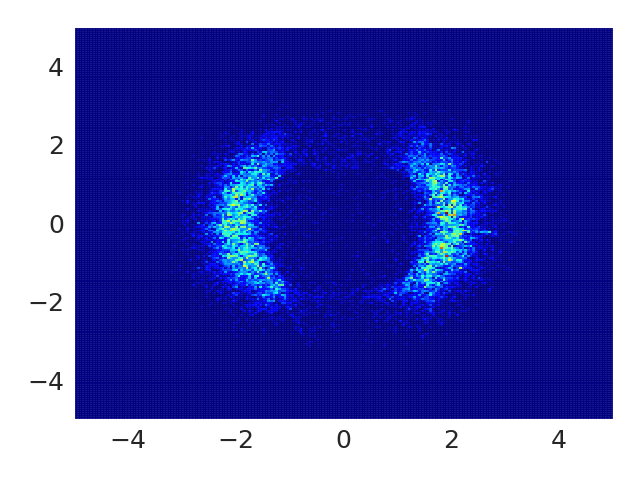}
    }
    \subfigure[Normalizing Flow]{
        \label{fig:moons-nf}
        \includegraphics[width=.23\linewidth]{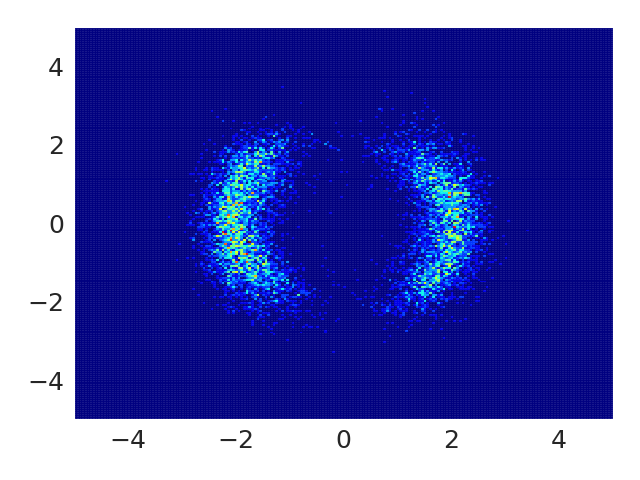}
    }
\vskip -0.1in

\caption{Twin Moons}   
\label{fig:additional-density-results-moons}
\vskip -0.05in
\end{figure*}

\begin{figure*}[ht!]
\centering
    \subfigure[Sinusoid Target]{
        \label{fig:sin-target}
        \includegraphics[width=.23\linewidth]{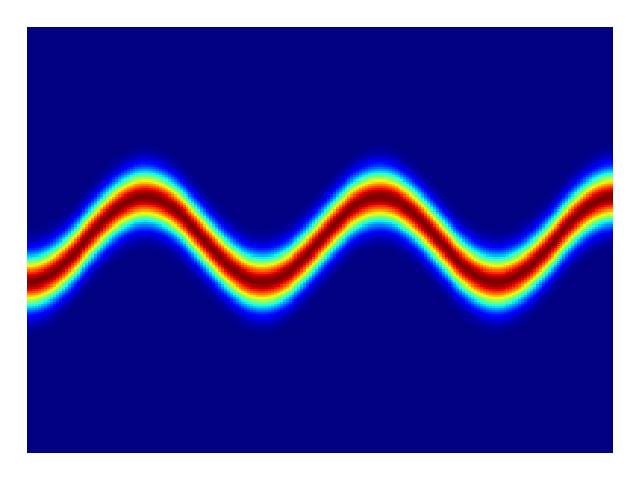}
    }
    \subfigure[GPVI]{
        \label{fig:sin-gpvi}
        \includegraphics[width=.23\linewidth]{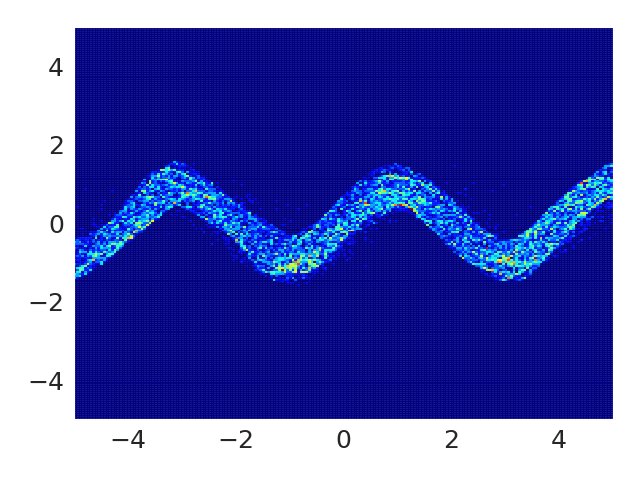}
    }
    \subfigure[Amortized SVGD]{
        \label{fig:sin-asvgd}
        \includegraphics[width=.23\linewidth]{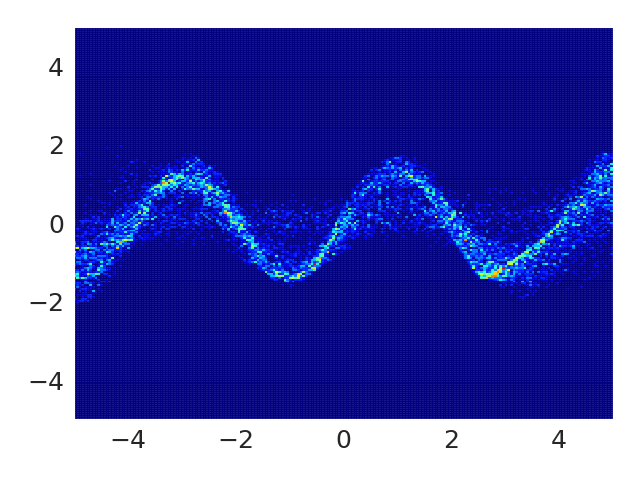}
    }
    \subfigure[Normalizing Flow]{
        \label{fig:sin-nf}
        \includegraphics[width=.23\linewidth]{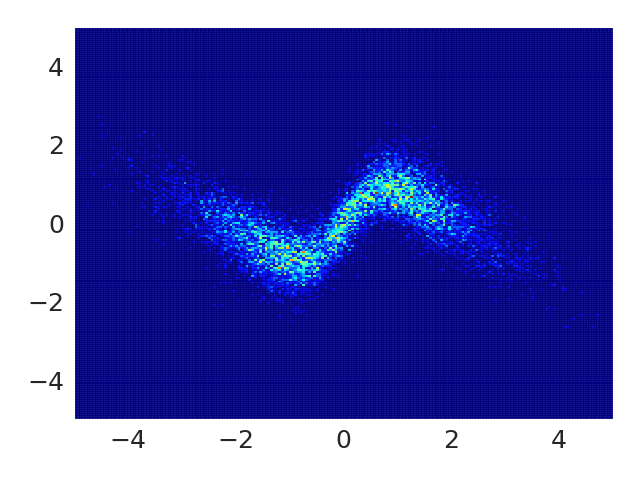}
    }
\vskip -0.1in
\caption{Sinusoid}   
\label{fig:additional-density-results-sin}
\vskip -0.05in
\end{figure*}

\begin{figure*}[ht!]
\centering
    \subfigure[Sin Bisect Target]{
        \label{fig:sin-target}
        \includegraphics[width=.23\linewidth]{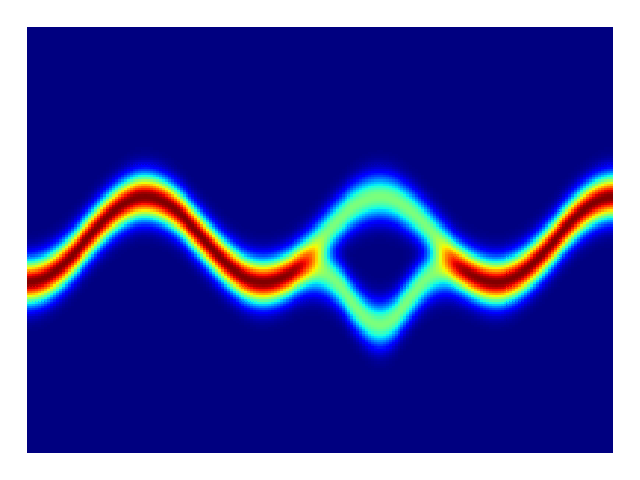}
    }
    \subfigure[GPVI]{
        \label{fig:sin-gpvi}
        \includegraphics[width=.23\linewidth]{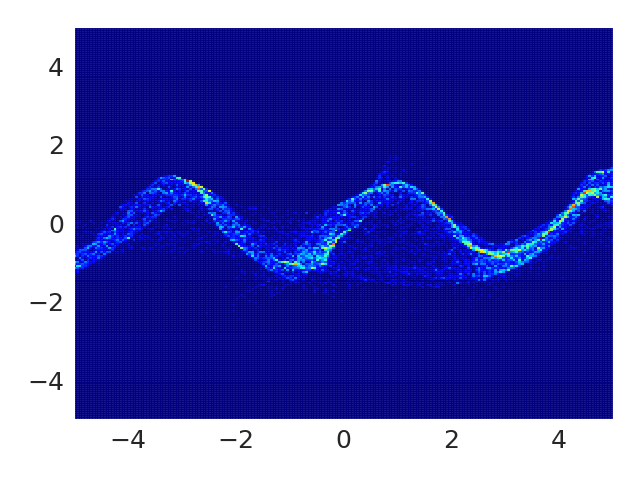}
    }
    \subfigure[Amortized SVGD]{
        \label{fig:sin-asvgd}
        \includegraphics[width=.23\linewidth]{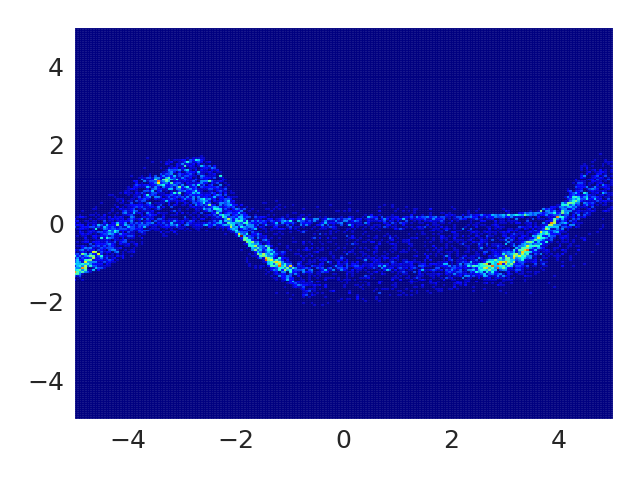}
    }
    \subfigure[Normalizing Flow]{
        \label{fig:sin-nf}
        \includegraphics[width=.23\linewidth]{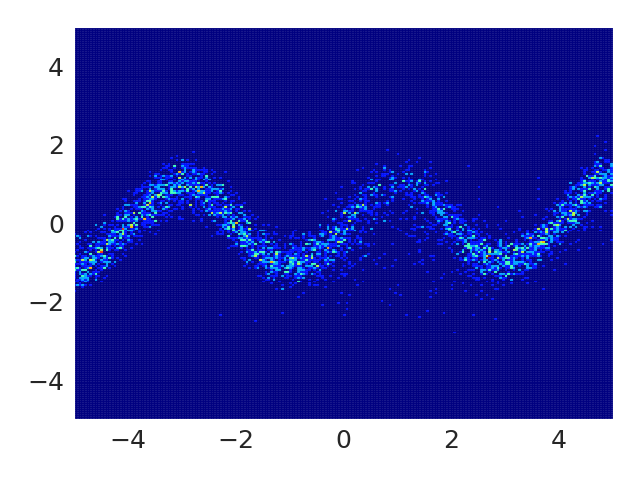}
    }
\vskip -0.1in
\caption{Sin Bisect}   
\label{fig:additional-density-results-sin-bisect}
\vskip -0.05in
\end{figure*}

\begin{figure*}[ht!]
\centering
    \subfigure[Sin Split Target]{
        \label{fig:sin-split-target}
        \includegraphics[width=.23\linewidth]{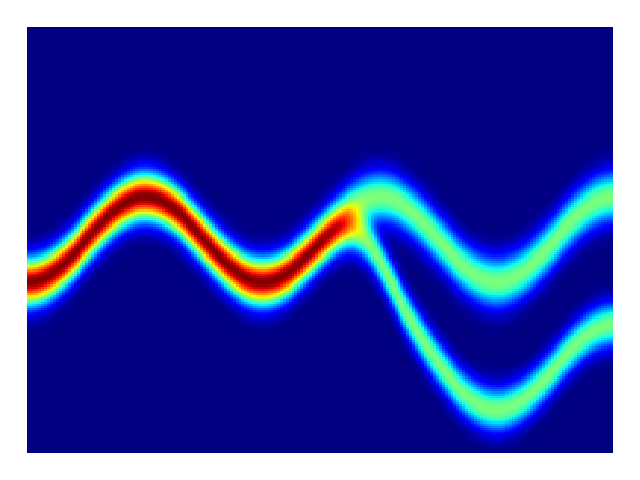}
    }
    \subfigure[GPVI]{
        \label{fig:sin-split-gpvi}
        \includegraphics[width=.23\linewidth]{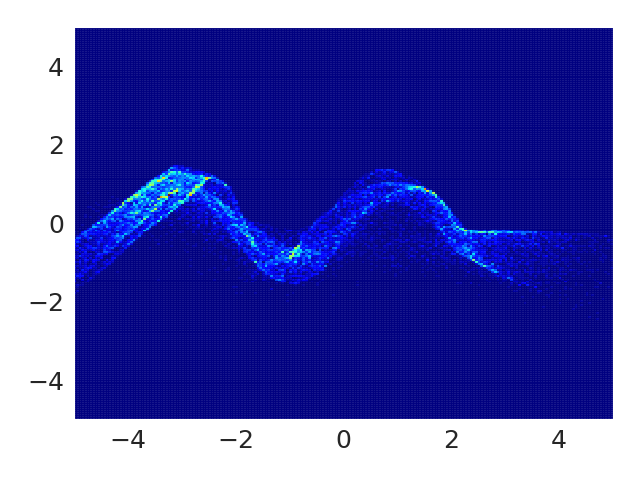}
    }
    \subfigure[Amortized SVGD]{
        \label{fig:sin-split-asvgd}
        \includegraphics[width=.23\linewidth]{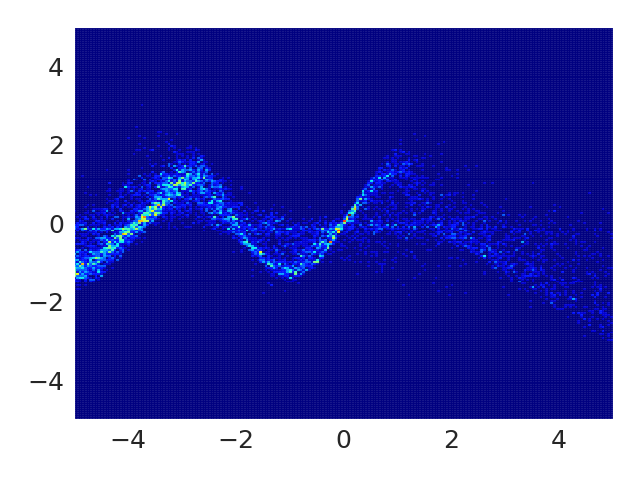}
    }
    \subfigure[Normalizing Flow]{
        \label{fig:sin-split-nf}
        \includegraphics[width=.23\linewidth]{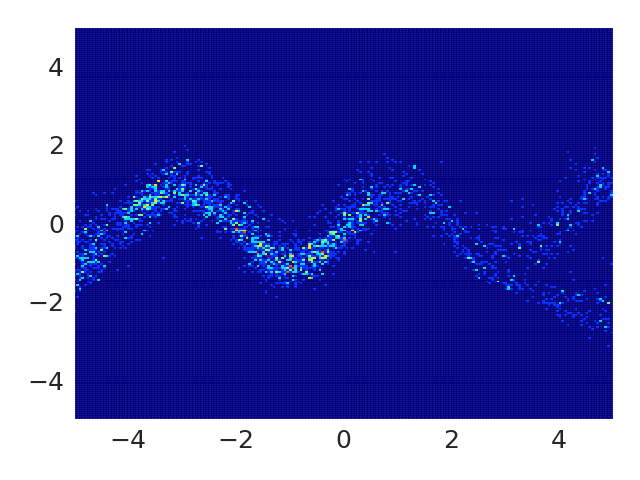}
    }
\vskip -0.1in
\caption{Sin Split}   
\label{fig:additional-density-results-sin-split}
\vskip -0.05in
\end{figure*}

We use the four non-Gaussian energy potentials defined in \citet{rezende2015variational}, and train GPVI as a sampler. In figures \ref{fig:additional-density-results-moons}-\ref{fig:additional-density-results-sin-split}, we see (from left to right) the target density, GPVI, amortized SVGD, and normalizing flows \citep{rezende2015variational}. We found that while our method has more parameters, we don't need nearly as deep a model as with normalizing flows. We used just 2 hidden layers for GPVI, while we needed a flow of depth 32 to get comparable performance. We trained each method for 200000 steps with a batch size of 100. To generate plots, we sampled 20000 points from each model. We detail the rest of the hyperparameters in section \ref{sec: hps}. In figures \ref{fig:additional-density-results-sin}, \ref{fig:additional-density-results-sin-bisect}, \ref{fig:additional-density-results-sin-split} we see that GPVI is able to capture the variance of the target distribution, while amortized SVGD samples mostly from the mean. For the Sin Bisect setting, GPVI is the only method that captures some of both halves of the middle section. But in Sin Split, only normalizing flows capture both halves of the split section. 

\subsection{Classification}
\begin{figure*}[ht!]
\centering
    \subfigure[GPVI]{
        \label{fig:gpvi_4class}
        \includegraphics[width=.30\linewidth]{figures/classification/rkhs-fg_4class.png}
    }
    \subfigure[SVGD]{
        \label{fig:svgd_4class}
        \includegraphics[width=.30\linewidth]{figures/classification/svgd_4class.png}
    }
    \subfigure[GFSF]{
        \label{fig:svgd_4class}
        \includegraphics[width=.30\linewidth]{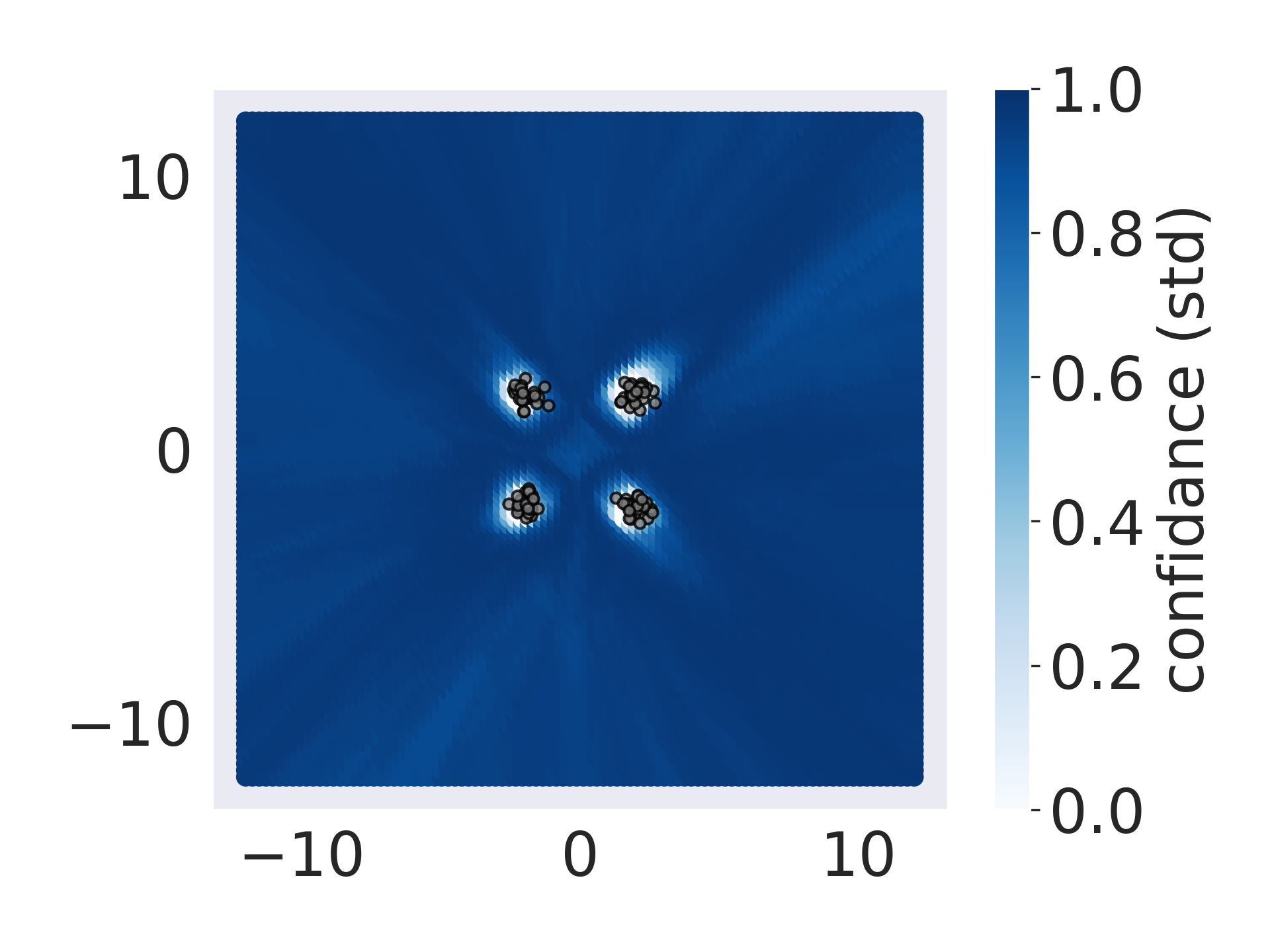}
    }
    \subfigure[KSD]{
        \label{fig:svgd_4class}
        \includegraphics[width=.30\linewidth]{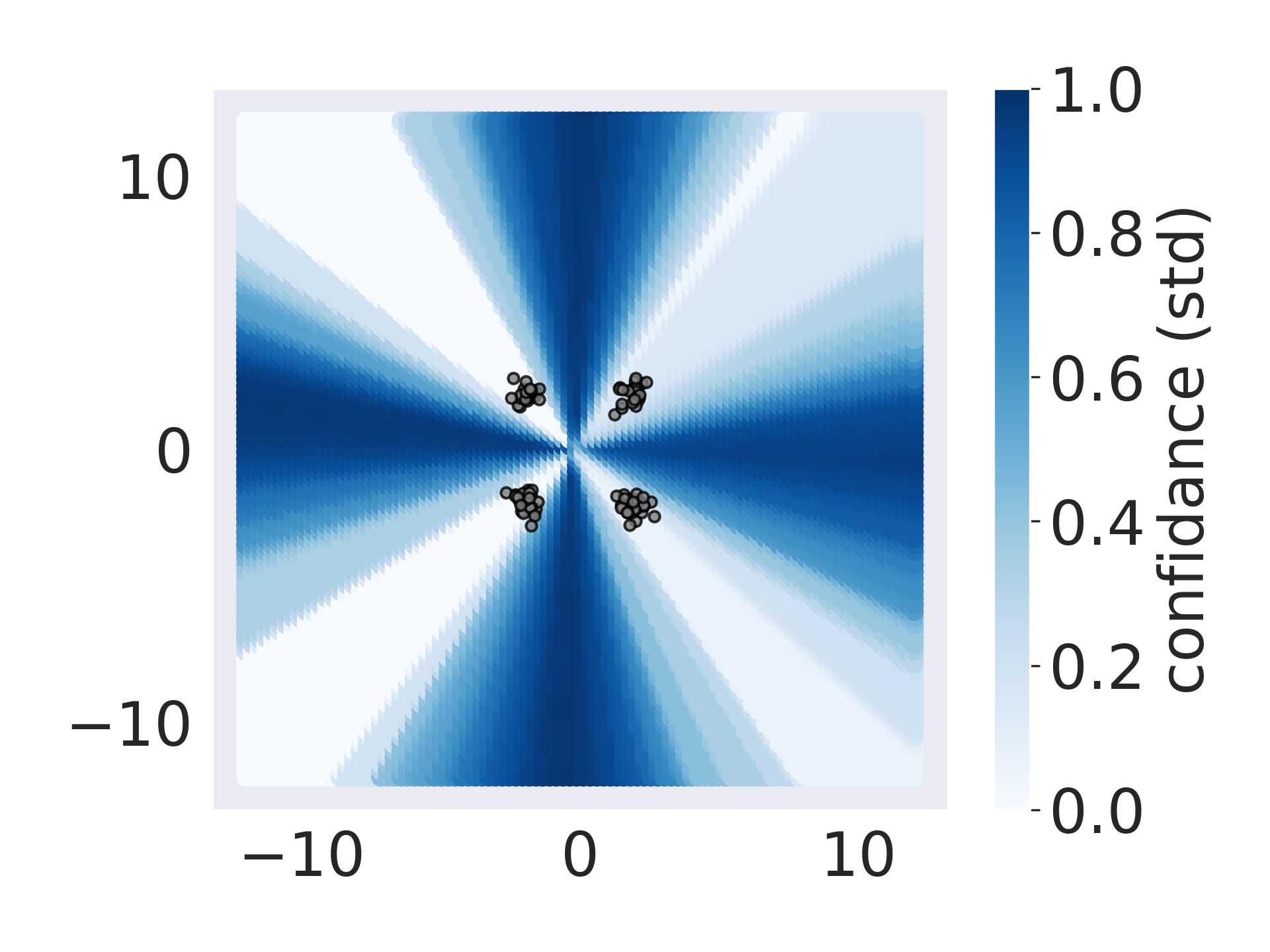}
    }
    \subfigure[HMC]{
        \label{fig:hmc_4class}
        \includegraphics[width=.30\linewidth]{figures/classification/hmc_4class.png}
    }
    \subfigure[Amortized SVGD]{
        \label{fig:asvgd_5class}
        \includegraphics[width=.30\linewidth]{figures/classification/asvgd_4class.png}
    }
    \subfigure[Amortized GFSF]{
        \label{fig:asvgd_5class}
        \includegraphics[width=.30\linewidth]{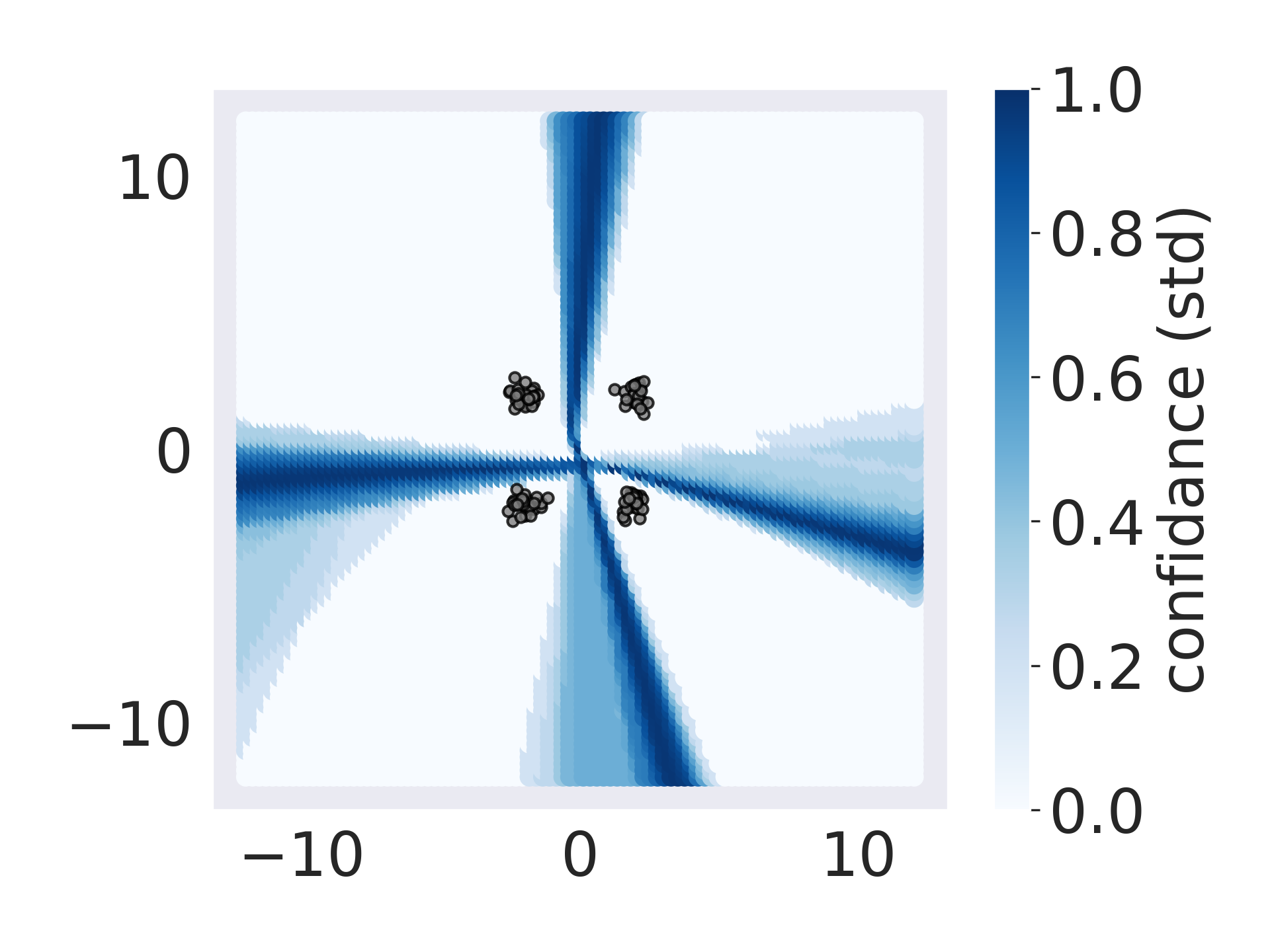}
    }
    \subfigure[Amortized KSD]{
        \label{fig:asvgd_5class}
        \includegraphics[width=.30\linewidth]{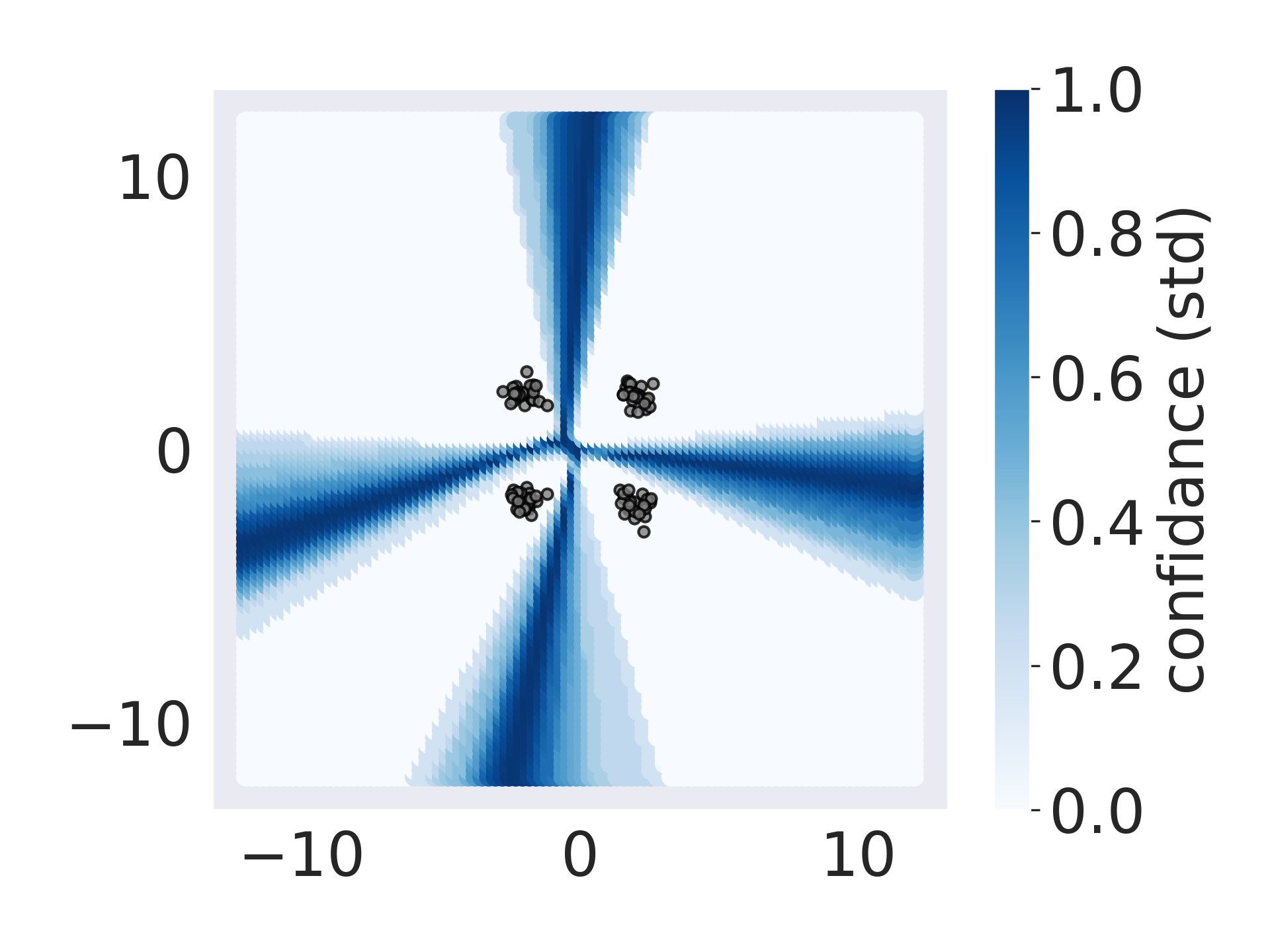}
    }
    \subfigure[MF-VI]{
        \label{fig:mfvi_4class}
        \includegraphics[width=.30\linewidth]{figures/classification/mf-vi_4class.png}
    }
    \subfigure[Deep Ensembles]{
        \label{fig:mfvi_4class}
        \includegraphics[width=.30\linewidth]{figures/classification/mle_4class.png}
    }
\vskip -0.1in
\caption{Predictive uncertainty of each method on the 4-class classification task, as measured by the standard deviation between predictions of sampled functions. }   
\label{fig:4class_classification-supp}
\vskip -0.05in
\end{figure*}

In figure \ref{fig:4class_classification-supp} we show the results of all evaluated methods on the four-class classification problem from section 4.3 of the main text. 

\begin{figure*}[ht!]
\centering
    \subfigure[GPVI]{
        \label{fig:gpvi_4class}
        \includegraphics[width=.30\linewidth]{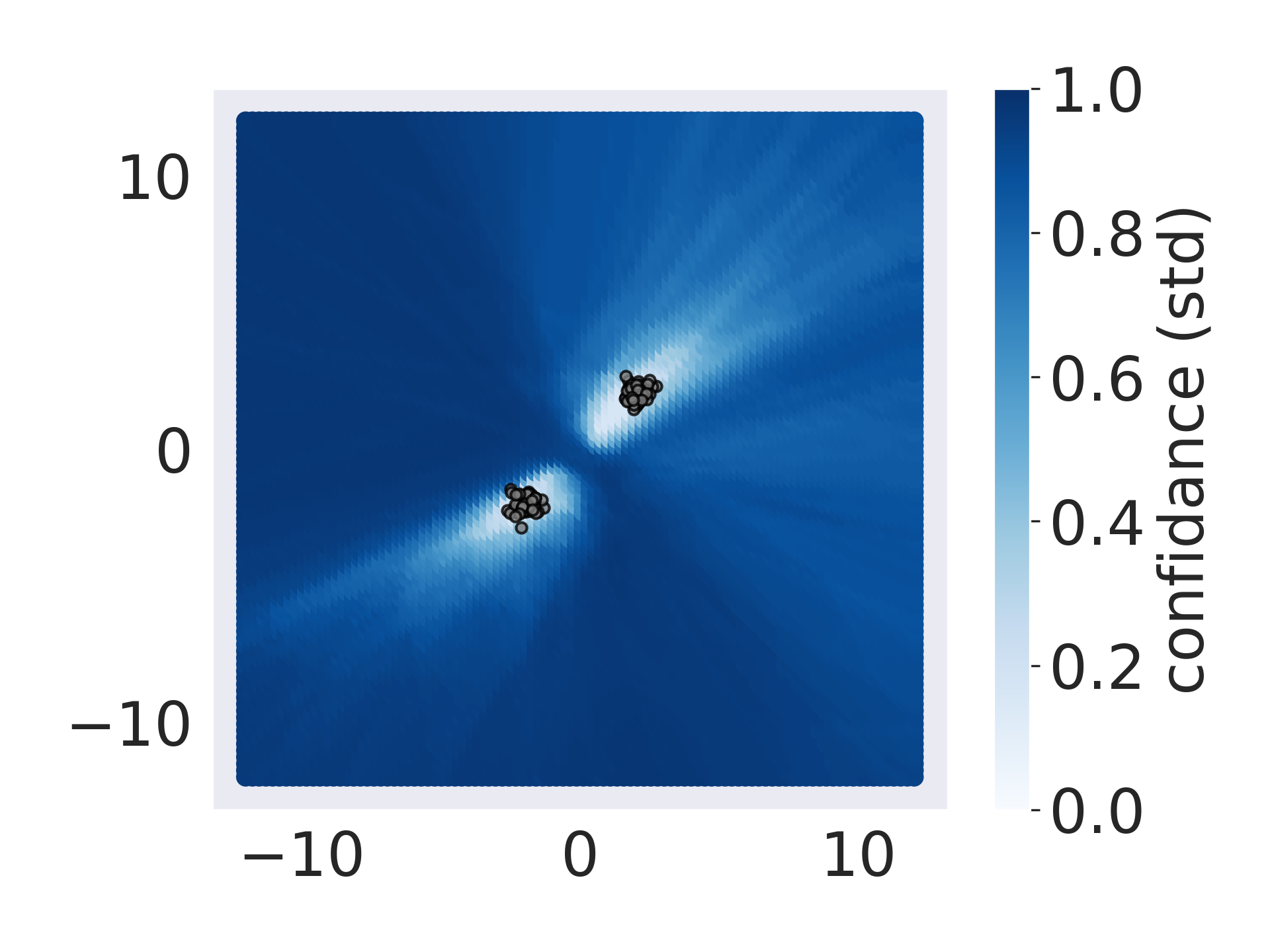}
    }
    \subfigure[SVGD]{
        \label{fig:svgd_4class}
        \includegraphics[width=.30\linewidth]{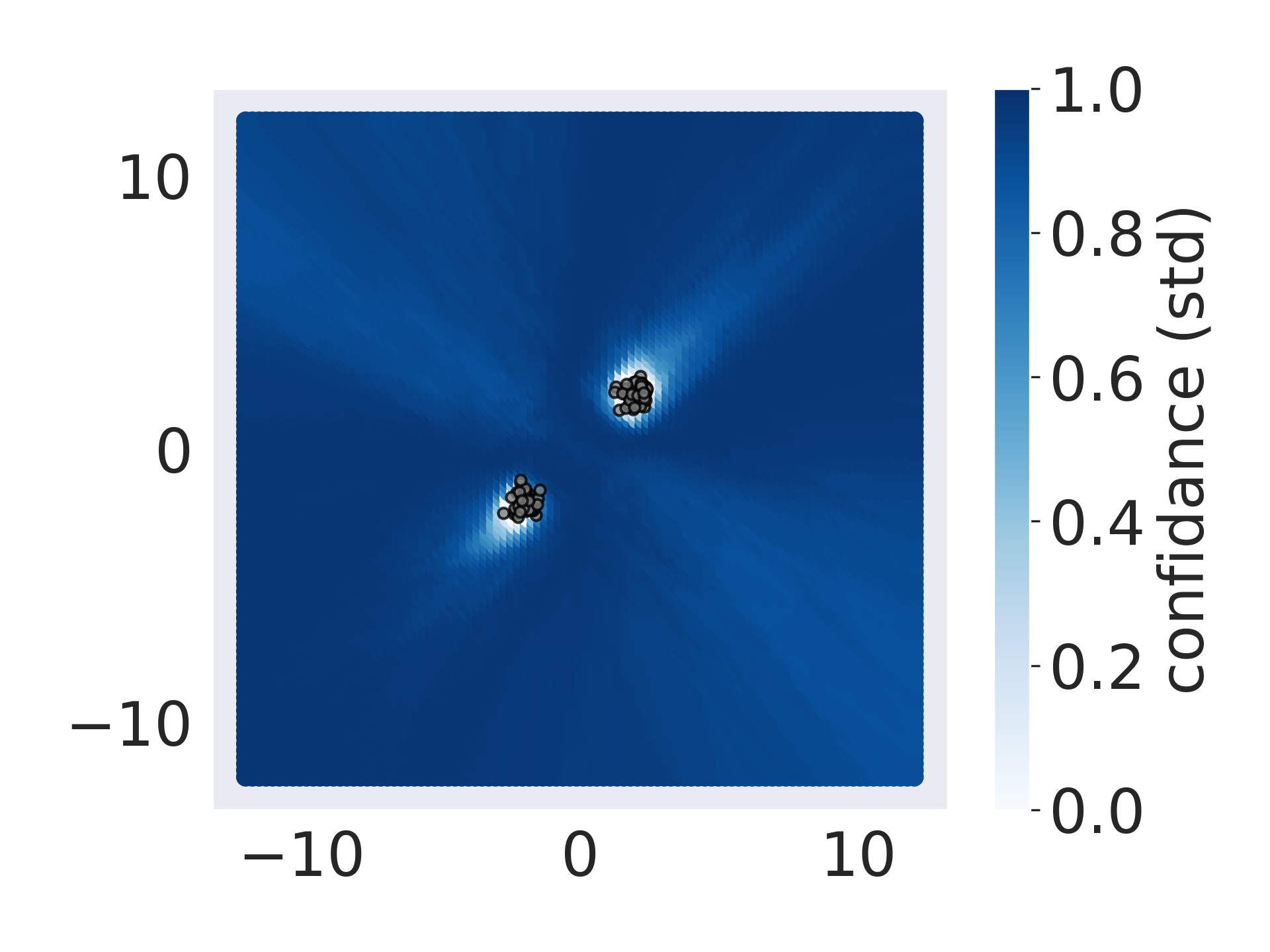}
    }
    \subfigure[GFSF]{
        \label{fig:svgd_4class}
        \includegraphics[width=.30\linewidth]{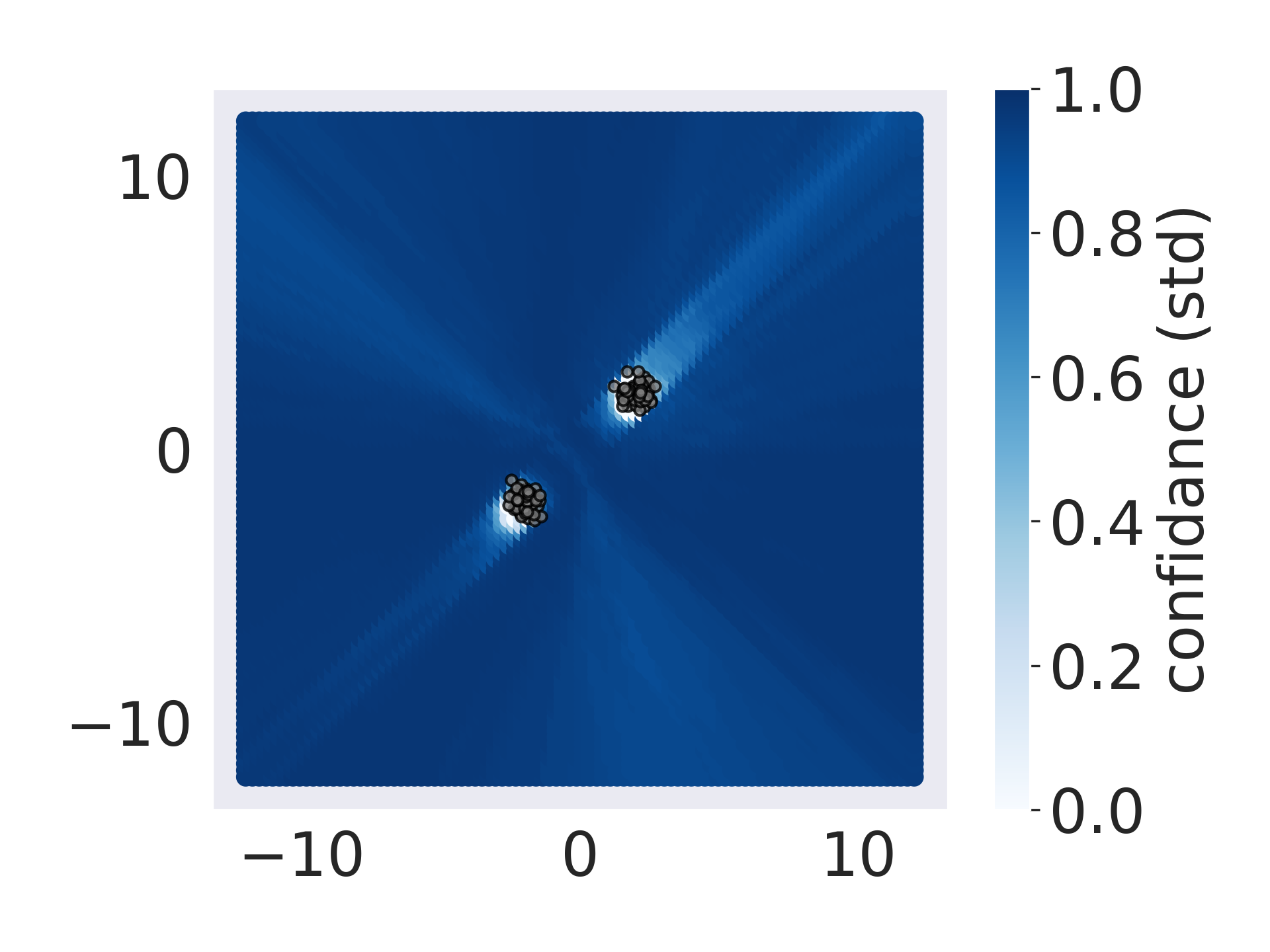}
    }
    \subfigure[KSD]{
        \label{fig:svgd_4class}
        \includegraphics[width=.30\linewidth]{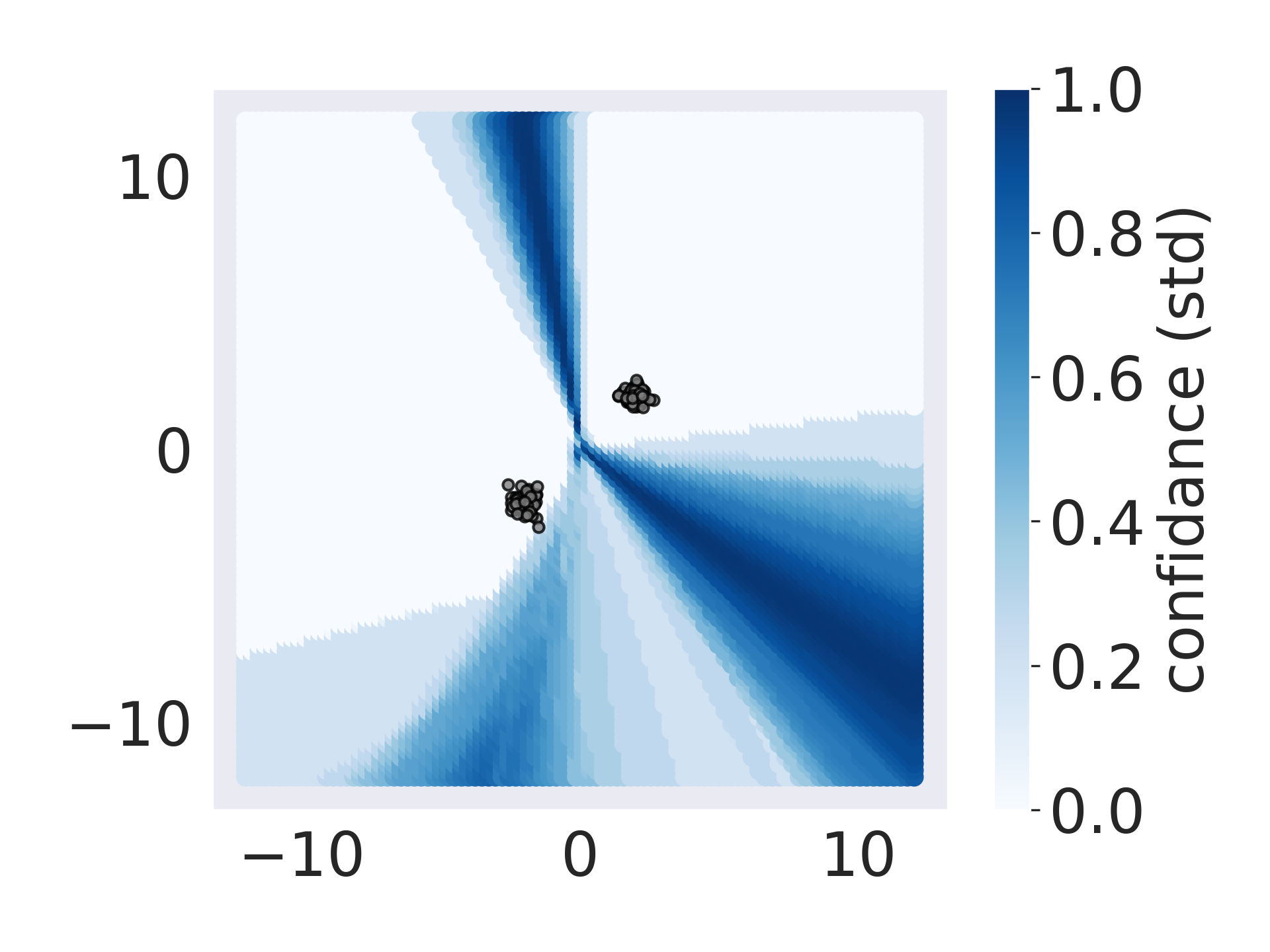}
    }
    \subfigure[HMC]{
        \label{fig:hmc_4class}
        \includegraphics[width=.30\linewidth]{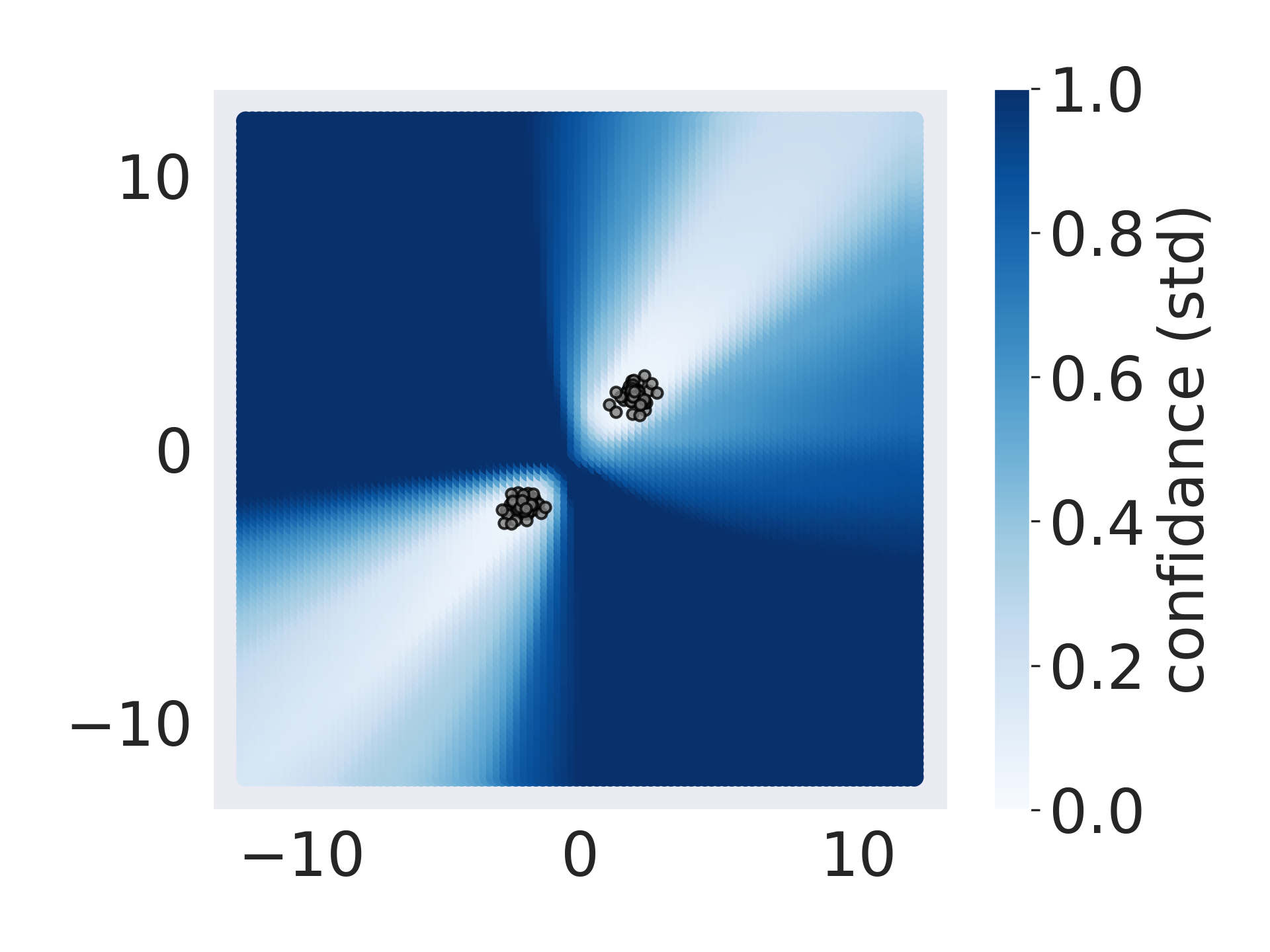}
    }
    \subfigure[Amortized SVGD]{
        \label{fig:asvgd_5class}
        \includegraphics[width=.30\linewidth]{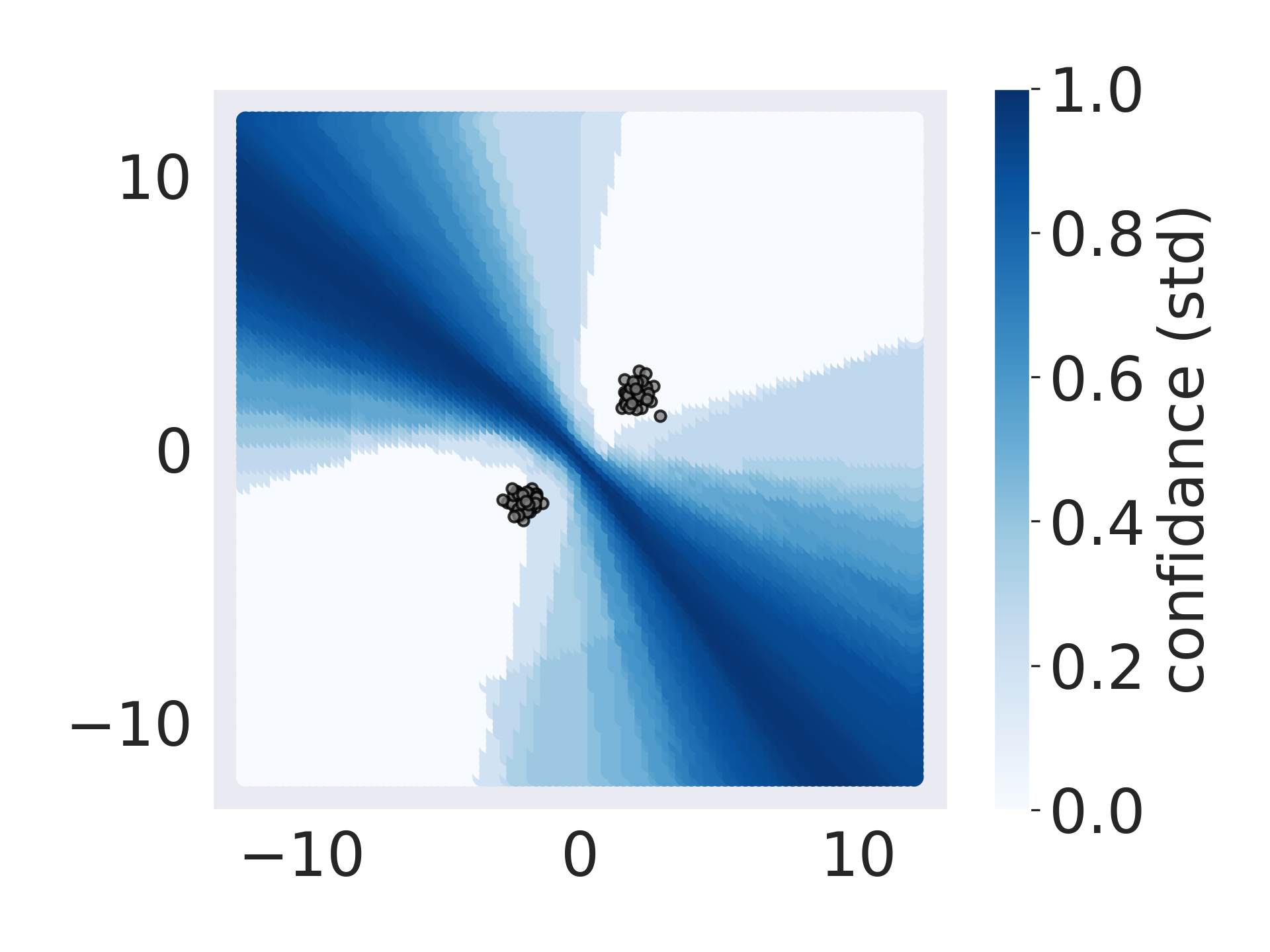}
    }
    \subfigure[Amortized GFSF]{
        \label{fig:asvgd_5class}
        \includegraphics[width=.30\linewidth]{figures/classification/2class/asvgd_2class.png}
    }
    \subfigure[Amortized KSD]{
        \label{fig:asvgd_5class}
        \includegraphics[width=.30\linewidth]{figures/classification/2class/asvgd_2class.png}
    }
    \subfigure[MF-VI]{
        \label{fig:mfvi_4class}
        \includegraphics[width=.30\linewidth]{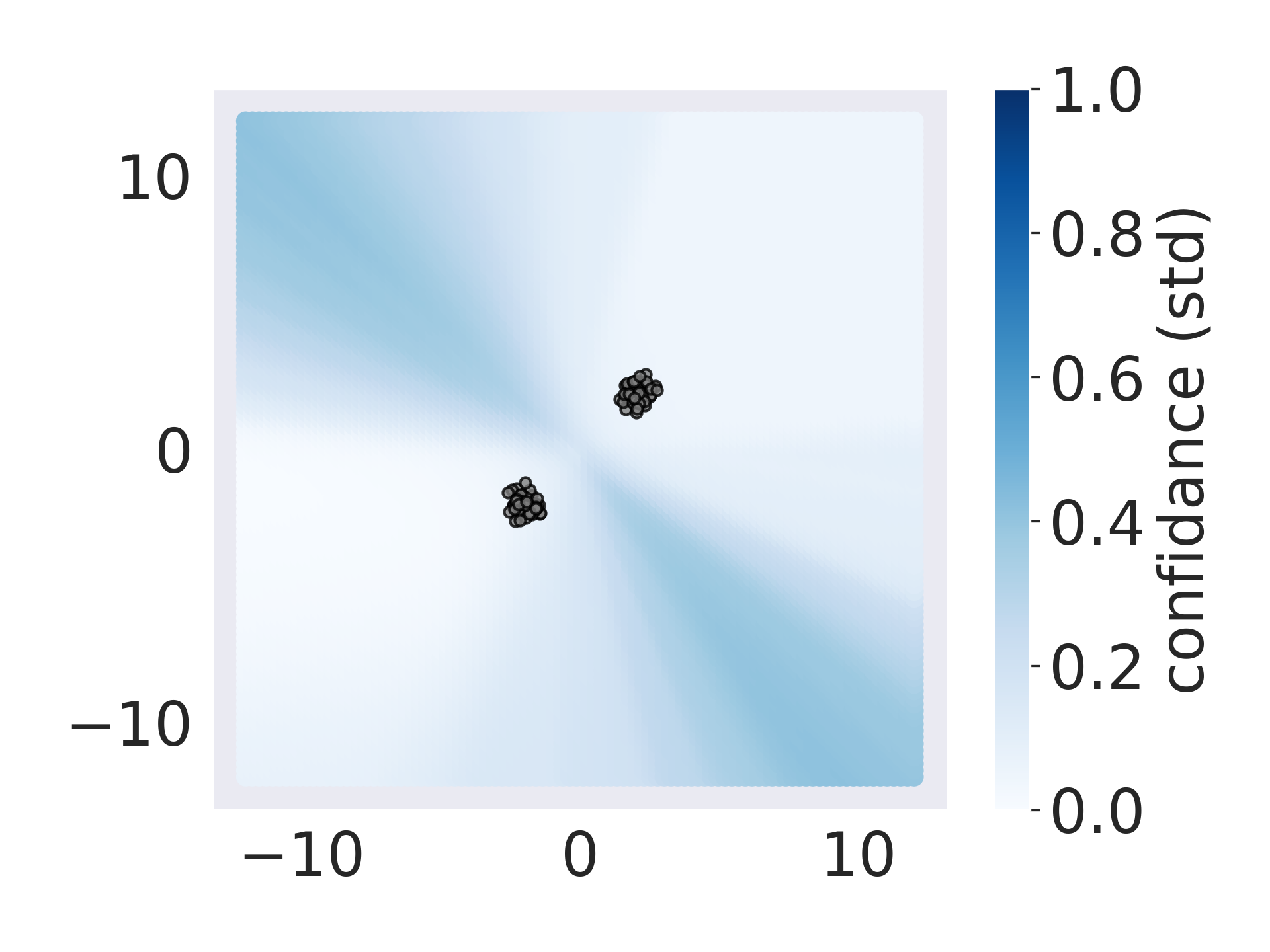}
    }
    \subfigure[Deep Ensembles]{
        \label{fig:mfvi_4class}
        \includegraphics[width=.30\linewidth]{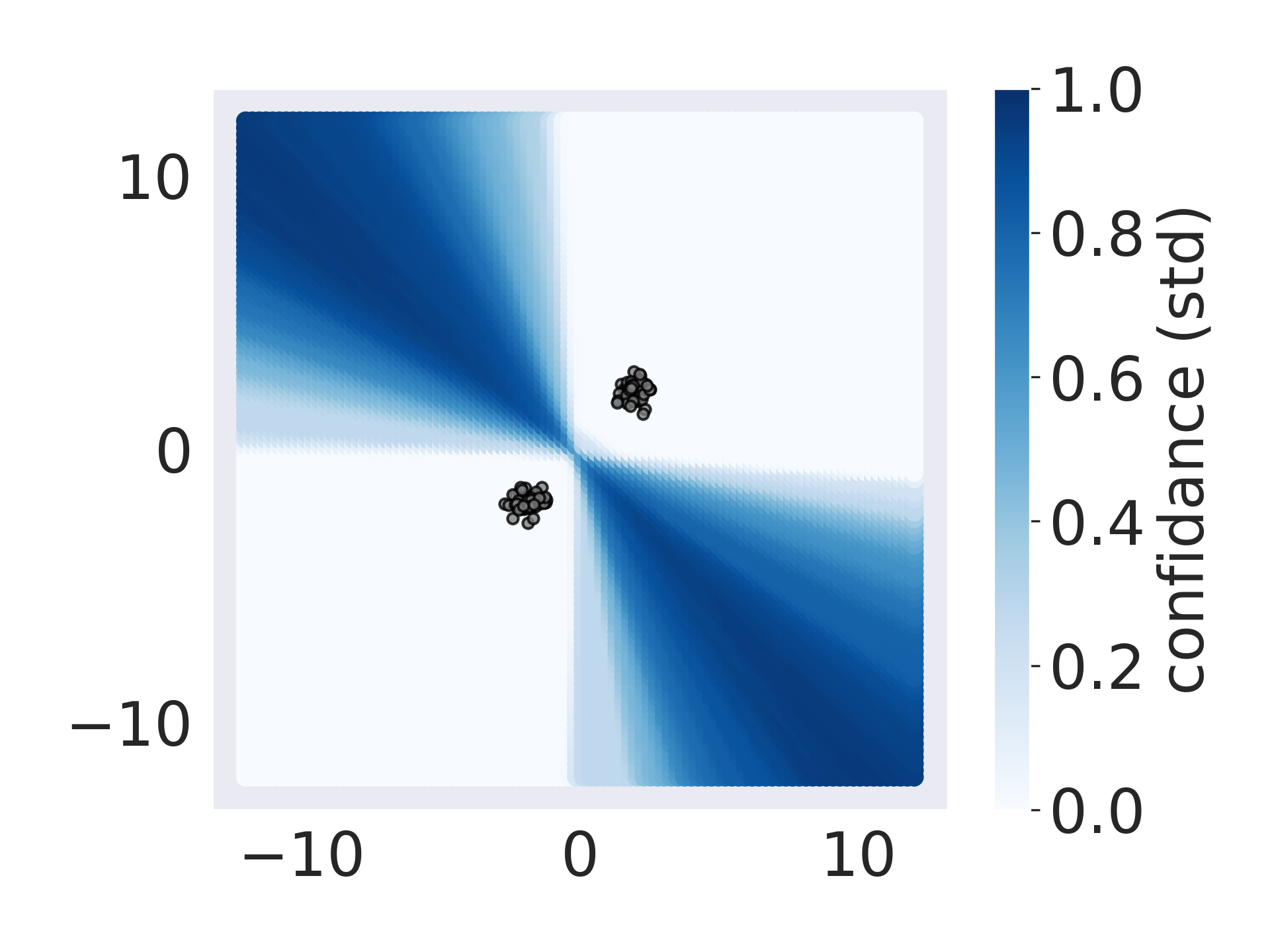}
    }
\vskip -0.1in
\caption{Predictive uncertainty of each method on the 2-class classification task, as measured by the standard deviation between predictions of sampled functions. }   
\label{fig:2class_classification}
\vskip -0.05in
\end{figure*}

In addition to the four-class classification problem we presented in the main paper, we show results on a simpler two class variant. 
In this setting, we generate data in the same way as in the four-class setting, but we use a mixture distribution with two components. 
Specifically, the mixture distribution is defined as $p(x) = \sum_{i=1}^2 \mathcal{N}(\mu_i, 0.3)$, with means $\mu_i \in \{(-2,-2), (2, 2)\}$. We assigned labels $y_i \in \{1, 2\}$ according to the index of the mixture component the samples were drawn from. We show the results of each method in figure \ref{fig:2class_classification}. We can see that GPVI again gives better uncertainty estimates than other sampling based approaches. Again, we do not know what the true posterior over classifications looks like, but GPVI and RKHS-based ParVI approaches give uncertainty estimates that closely match our intuition for this problem. 

\section{Experimental Details}
Here we provide some additional details regarding the setup and reporting of our experiments and their results.

\par\noindent\textbf{HMC Details}
We use the same settings for HMC for the Bayesian linear regression and 2/4 class classification tasks. We sampled the momentum from a standard normal distribution, and used 25 leap-frog steps per sample. We ran each experiment for 25K steps in total, with a burn-in of 20K steps, and a step size of $0.0005$. We thinned each chain after burn-in, and tuned the
number of leapfrog steps and step-size for each experiment.
To check for convergence, we checked that the means of each chain were similar, indicating mixing. 

\par\noindent\textbf{1D Regression}
For our 1D regression task (figure 1 of the main paper), we generate a dataset of 80 samples $X$ with targets $Y$. We draw 76 samples of $X$ uniformly from $[-6, -2] \cup [2, 6]$, and draw the 4 remaining samples from $[-2, 2]$. The targets $Y$ are computed as $Y=-(1+X) \sin(1.2X) + \epsilon$, where $\epsilon \sim \mathcal{N}(0, 0.04)$. 
For all methods we use 100 posterior samples, and train for 50K iterations. For GPVI and amortized SVGD we use a generator with two linear layers $[32, 32]$ and Gaussian input noise of the same dimension. 

\par\noindent\textbf{Density Estimation}\\
In the density estimation setting (section 4.1 of the main text), we randomly generate $2d$ and $5d$ target covariance matrices. Specifically, the target covariance we wish to fit is computed as $\Sigma^* = \Sigma\Sigma^T, \Sigma \sim \mathcal{N}(0, I_d), d \in \{2, 5\}$. We use an MLP with 1 hidden layer of width 2 to approximately sample from a unimodal Gaussian distribution with covariance $\Sigma^*$. Given Gaussian input noise, the variance of the output distribution of the linear generator is computed as $W^TW$, which should match $\Sigma^*$ after training. 

\par\noindent\textbf{Open Category Tasks}\\
For the open-category tasks, we computed two metrics to evaluate each method: AUROC (AUC) and ECE. In this context, AUC measures how well a binary classifier can discriminate between predictions made on inlier vs outlier test inputs. An AUC score of $1.0$ indicates that the predictions made by a model are perfectly separable, meaning that we could set a threshold on predicted probabilities to detect every outlier test input. In practice we first make predictions on the inlier and outlier test sets, then compute the variance of the predictions over the sampled predictors. The AUC score is then computed against the predictive variance using the scikit-learn library.    

Expected Calibration Error (ECE) measures how well calibrated a model's predictions are. Given predictions on inlier test inputs, ECE partitions predictions into $M=15$ equally sized bins according to their confidence. For each bin $b_m$ we compute the difference between their accuracy $1/|b_m| \sum_{x_i \in b_m} (\hat{y} - y)$ and confidence $1/b_m \sum_{x_i \in m} \hat{p}$. Where $\hat{p}, \hat{y}$ are predicted probabilities and associated label with respect to a data-point $x_i$ and true label $y$. This difference in expected accuracy and confidence between each bin represents the bin's calibration error. 

For all experiments we report an average of three runs using the hyperparameter settings given in section \ref{sec: hps}. 

\section{Hyperparameter Settings} 
\label{sec: hps}
In tables \ref{tab:density-energy-params} - \ref{tab:cifar10-params} we detail the hyperparameters chosen for each method in each experimental setting. We refer to the sampler network as $\vec{f}$, and the predicting classifier as $\vec{g}$. We generally use the same structure of $\vec{f}$ and input noise for GPVI and amortized ParVI methods. 

\begin{longtable}[ht]{rr l }
    \toprule
    \multicolumn{2}{ c }{Density Estimation (energy potentials)}  \\
    \midrule
    \multicolumn{2}{r}{Hyperparameter} &  Value\\
    \midrule
    \multicolumn{2}{r}{\textit{Common}} & \\
    & Posterior Samples & $20000$ \\
    & Learning Rate & $1e-4$\\
    & $\vec{f}$ Hidden Layers & $2$ \\
    & $\vec{f}$ Hidden Width & $[500, 500]$ \\
    & $\vec{f}$ Input Noise Stdev & $\sigma \in \{1.0, 2.0, 6.0\}$ \\

    & Training Steps & $200e3$\\
    & Minibatch Size & $100$\\
    \midrule
    \multicolumn{2}{r}{\textit{GPVI}} & \\
    & Optimizer & Adam \\
    & $\vec{h}_\eta$ Hidden Width   & $500$\\
    & $\vec{h}_\eta$ Hidden Layers & $3$\\
    & $\vec{h}_\eta$ Learning Rate & $1e-4$\\
    \midrule
    \multicolumn{2}{r}{\textit{Normalizing Flow}} & \\
    & Optimizer & RMSProp \\
    & Flow Length & $32$ \\
    & Weight Decay & $1e-3$ \\
    & Base Dist Stdev & $\sigma \in \{1.0, 2.0, 6.0\}$ \\
    & Flow Architecture & Planar Flow \\
    \bottomrule
    \caption{Hyperparameters for density estimation of energy potentials.}.
   \label{tab:density-energy-params}
\end{longtable}

\begin{longtable}[ht]{rr l }
    \toprule
    \multicolumn{2}{ c }{Bayesian Linear Regression}  \\
    \midrule
    \multicolumn{2}{r}{Hyperparameter} &  Value\\
    \midrule
    \multicolumn{2}{r}{\textit{Common}} & \\
    & Optimizer (all) & Adam \\
    & Posterior Samples & $100$ \\
    & Learning Rate & $1e-3$\\
    & $\vec{f}$ Hidden Layers & None \\
    & $\vec{f}$ Input Noise $\vec{z}$ & $\mathcal{N}(0, I_{3})$\\
    & $\vec{g}$ Hidden Layers & None \\
    & Training Steps & $50e3$\\
    & Minibatch Size & $10$\\\\\\
    \midrule
    \multicolumn{2}{r}{\textit{GPVI}} & \\
    & $\vec{h}_\eta$ Hidden Width   & $10$\\
    & $\vec{h}_\eta$ Hidden Layers & $3$\\
    & $\vec{h}_\eta$ Learning Rate & $1e-4$\\
    \midrule
    \multicolumn{2}{r}{\textit{KSD (Amortized and ParVI)}} & \\
    & Critic Hidden Layers & $1$ \\
    & Critic Hidden Width & $100$ \\
    & Critic Learning Rate & $1e-3$ \\
    & Critic $L_2$ Weight & $10$ \\
    \midrule
    \multicolumn{2}{r}{\textit{MF-VI (Bayes by Backprop)}} & \\
    & Weight Prior                 & Scale Mixture\\
    & Mixture Weight $\pi$         & $0.5$\\
    & $\sigma_1$                   & $1.0$\\
    & $\sigma_2$                   & $0$\\
    \bottomrule
    \caption{Hyperparameters for Bayesian linear regression task}.
   \label{tab:blr-params}
\end{longtable}

\begin{longtable}[ht]{rr l }
    \toprule
    \multicolumn{2}{c}{4/2-class Classification} \\
    \midrule
    \multicolumn{2}{r}{Hyperparameter} &  Value\\
    \midrule
    \multicolumn{2}{r}{\textit{Common}} & \\
    & Optimizer (all) & Adam \\
    & Posterior Samples & $100$ \\
    & Learning Rate & $1e-3$\\
    & Training Steps & 500e3\\
    & Minibatch Size & 100\\
    & $\vec{f}$ Hidden Layers & 2 \\
    & $\vec{f}$ Hidden Width & 64 \\
    & $\vec{f}$ Nonlinearity & ReLU \\
    & $\vec{f}$ Input Noise $\vec{z}$ & $\mathcal{N}(0, I_{64})$\\
    & $\vec{g}$ Hidden Layers & 2 \\
    & $\vec{g}$ Hidden Width & 10 \\
    & $\vec{g}$ Nonlinearity & ReLU \\
    \midrule
    \multicolumn{2}{r}{\textit{GPVI}} & \\
    & $\vec{h}_\eta$ Hidden Width   & $544$\\
    & $\vec{h}_\eta$ Hidden Layers & $3$\\
    & $\vec{h}_\eta$ Learning Rate & $1e-4$\\
    \midrule
    \multicolumn{2}{r}{\textit{KSD (Amortized and ParVI)}} & \\
    & Critic Hidden Layers & $2$ \\
    & Critic Hidden Width & $100$ \\
    & Critic Learning Rate & $1e-3$ \\
    & Critic $L_2$ Weight & $10$ \\
    \midrule
    \multicolumn{2}{r}{\textit{MF-VI (Bayes by Backprop)}} & \\
    & Weight Prior                 & Scale Mixture\\
    & Mixture Weight $\pi$         & $0.5$\\
    & $\sigma_1$                   & $1.0$\\
    & $\sigma_2$                   & $\exp{(-6)}$\\
    \bottomrule
    \caption{Hyperparameters for 2/4 class classification task}.
    \label{tab:classification-params}
\end{longtable}

\begin{longtable}[ht]{rr l }
    \toprule
    \multicolumn{2}{r}{\emph{Open-Category (MNIST)}}& \\
    \midrule
    \multicolumn{2}{r}{\textit{Common}} & \\
    & Optimizer (all) & Adam \\
    & Posterior Samples & $10$ \\
    & Training Epochs & $100$\\
    & Minibatch Size & $50$\\
    & $\vec{f}$ Learning Rate & $1e-5$ \\
    & $\vec{f}$ Hidden Layers & $3$ \\
    & $\vec{f}$ Hidden Width & $[256, 512, 1024]$ \\
    & $\vec{f}$ Nonlinearity & ReLU \\
    & $\vec{f}$ Input Noise $\vec{z}$ & $\mathcal{N}(0, I_{256})$\\
    & $\vec{g}$ Learning Rate & $1e-5$ \\
    & $\vec{g}$ Architecture & LeNet-5 \\
    & $\vec{g}$ Nonlinearity & ReLU \\
    \midrule
    \multicolumn{2}{r}{\textit{GPVI}} & \\
    & $\vec{h}_\eta$ Hidden Width   & $512$\\
    & $\vec{h}_\eta$ Hidden Layers & $3$\\
    & $\vec{h}_\eta$ Learning Rate & $1e-4$\\
    \midrule
    \multicolumn{2}{r}{\textit{KSD (Amortized and ParVI)}} & \\
    & Critic Hidden Layers & $2$ \\
    & Critic Hidden Width & $512$ \\
    & Critic Learning Rate & $1e-4$ \\
    & Critic $L_2$ Weight & $1.0$ \\
    \midrule
    \multicolumn{2}{r}{\textit{MF-VI (Bayes by Backprop)}} & \\
    & Weight Prior                 & Scale Mixture\\
    & Mixture Weight $\pi$         & $0.5$\\
    & $\sigma_1$                   & $1.0$\\
    & $\sigma_2$                   & $\exp{(-6)}$\\
    \bottomrule
   \caption{Hyperparameters for MNIST open category task}.
   \label{tab:MNIST-params}
\end{longtable}

\begin{longtable}[ht]{rr l }
    \toprule
    \multicolumn{2}{r}{\emph{Open-Category (CIFAR-10)}}& \\
    \midrule
    \multicolumn{2}{r}{\textit{Common}} & \\
    & Optimizer (all) & Adam \\
    & Posterior Samples & $10$ \\
    & Training Epochs & $200$\\
    & Minibatch Size & $50$\\
    & $\vec{f}$ Learning Rate & $1e-5$ \\
    & $\vec{f}$ Hidden Layers & $3$ \\
    & $\vec{f}$ Hidden Width & $[400, 600, 1000]$ \\
    & $\vec{f}$ Nonlinearity & ReLU \\
    & $\vec{f}$ Input Noise $\vec{z}$ & $\mathcal{N}(0, I_{400})$\\
    & $\vec{g}$ Hidden Layers & $3$ conv, $2$ linear \\
    & $\vec{g}$ Hidden Width & $[32, 64, 64, 128, 10]$ \\
    & $\vec{g}$ Nonlinearity & ReLU \\
    \midrule
    \multicolumn{2}{r}{\textit{GPVI}} & \\
    & $\vec{h}_\eta$ Hidden Width   & $512$\\
    & $\vec{h}_\eta$ Hidden Layers & $3$\\
    & $\vec{h}_\eta$ Learning Rate & $1e-4$\\\\
    \midrule
    \multicolumn{2}{r}{\textit{KSD (Amortized and ParVI)}} & \\
    & Critic Hidden Layers & $2$ \\
    & Critic Hidden Width & $512$ \\
    & Critic Learning Rate & $1e-4$ \\
    & Critic $L_2$ Weight & $1.0$ \\
    \midrule
    \multicolumn{2}{r}{\textit{MF-VI (Bayes by Backprop)}} & \\
    & Weight Prior                 & Scale Mixture\\
    & Mixture Weight $\pi$         & $0.5$\\
    & $\sigma_1$                   & $1.0$\\
    & $\sigma_2$                   & $\exp{(-6)}$\\
    \bottomrule
   \caption{Hyperparameters for CIFAR-10 open category task}.
   \label{tab:cifar10-params}
\end{longtable}


\end{document}